\documentclass[letterpaper, 10 pt, conference]{ieeeconf}  
\IEEEoverridecommandlockouts  
\usepackage{graphics} 
\usepackage{epsfig} 
\usepackage{times} 
\usepackage{amsmath} 
\usepackage{amssymb}  
\let\labelindent\relax

\usepackage{graphicx}
\usepackage{booktabs}
\usepackage[section]{placeins}
\usepackage{cleveref}

\crefname{algocf}{alg.}{algs.}
\Crefname{equation}{Eq.}{Eqs.}
\creflabelformat{equation}{#2\textup{#1}#3}
\Crefname{figure}{Fig.}{Figs.}

\usepackage[noadjust]{cite}
\usepackage{multicol}
\usepackage{setspace}
\usepackage{thmtools,thm-restate}
\usepackage{xspace}
\usepackage{bbm}
\usepackage{color}
\usepackage{mathtools}
\usepackage[noend, ruled, linesnumbered]{algorithm2e}
\usepackage[noend]{algpseudocode}
\usepackage{subcaption}
\usepackage{multirow}
\usepackage{tabulary}
\usepackage{comment}

\graphicspath{{./figs/}}
\newcommand{\Cspace}{\ensuremath{\calX}\xspace}
\newcommand{\Cfree}{\ensuremath{\calX_{\rm free}}\xspace}
\newcommand{\Cobs}{\ensuremath{\calX_{\rm obs}}\xspace}
\newcommand{\Cinf}{\ensuremath{\calX_{\rm IS}}\xspace}

\newcommand{\vs}{\ensuremath{v_{\rm s}}}
\newcommand{\vg}{\ensuremath{v_{\rm t}}}
\newcommand{\xs}{\ensuremath{x_{\rm s}}}
\newcommand{\xg}{\ensuremath{x_{\rm t}}}

\newcommand{\graph}{\ensuremath{\calG}}

\newcommand{\searchTree}[0]{\ensuremath{\calT}}

\newcommand{\ellipse}{\ensuremath{\calE}}

\newcommand{\sampleSet}{\ensuremath{\calS}\xspace}

\newcommand{\ellipseinf}[0]{\ensuremath \ellipse_{\rm IS} \xspace}
\newcommand{\ellipselocal}[0]{\ensuremath \ellipse_{\rm LS} \xspace}

\newcommand{\Path}[0]{\ensuremath \xi \xspace}
\newcommand{\PathSet}[0]{\ensuremath \Xi \xspace}

\newcommand\algname[2]{\texttt{#1}{#2}\xspace}

\newcommand{\guild}{\algname{GuILD}{}}

\newcommand{\uniformguild}{\algname{Uniform}{}}
\newcommand{\greedyguild}{\algname{Greedy}{}}
\newcommand{\banditguild}{\algname{Bandit}{}}

\newcommand{\forestenv}{\algname{Forest}{}}
\newcommand{\twowallforestenv}{\algname{TwoWall}{}}
\newcommand{\lforestenv}{\algname{Trap}{}}
\newcommand{\carenv}{\algname{SE2Maze}{}}
\newcommand{\herbenv}{\algname{HERB Bookshelf}{}}

\newcommand{\BeaconSelector}[0]{\textsc{BeaconSelector}\xspace}
\newcommand{\beaconSet}{\ensuremath{\calB}}
\DeclareMathOperator*{\argmin}{arg\,min}
\DeclareMathOperator*{\argmax}{arg\,max}

\newcommand{\calB}{\ensuremath{\mathcal{B}}\xspace}

\newcommand{\calE}{\ensuremath{\mathcal{E}}\xspace}
\newcommand{\calG}{\ensuremath{\mathcal{G}}\xspace}
\newcommand{\calK}{\ensuremath{\mathcal{K}}\xspace}

\newcommand{\calS}{\ensuremath{\mathcal{S}}\xspace}
\newcommand{\calT}{\ensuremath{\mathcal{T}}\xspace}
\newcommand{\calU}{\ensuremath{\mathcal{U}}\xspace}

\newcommand{\calX}{\ensuremath{\mathcal{X}}\xspace}

\newcommand{\R}{\mathbb{R}} 

\newtheorem{theorem}{Theorem}[section]

\newcommand{\ignore}[1]{}

\usepackage{enumitem}
\newlist{hypothesis}{enumerate}{1}
\setlist[hypothesis]{resume, label=\textbf{H\arabic*}, labelindent=\parindent, leftmargin=*}

\definecolor{orange}{rgb}{1,0.5,0}
\definecolor{internationalorange}{rgb}{1.0, 0.31, 0.0}

\newcommand{\xxnote}[3]{}
\ifx\hidenotes\undefined
  \usepackage{color}
  \renewcommand{\xxnote}[3]{\color{#2}{#1: #3}}
\fi

\title{\LARGE \bf
Guided Incremental Local Densification for \\ Accelerated Sampling-based Motion Planning
}

\author{Aditya Mandalika$^{*1}$, Rosario Scalise$^{*1}$, Brian Hou$^{*1}$, Sanjiban Choudhury$^{2}$ and 
Siddhartha S. Srinivasa$^{1}$
\thanks{$^{*}$ All three authors contributed equally.}
\thanks{$^{1}$Paul G. Allen School of Computer Science and Engineering, University of Washington 
         \texttt{\small \{adityavk, rosario, bhou, siddh\}@cs.washington.edu}}%
\thanks{$^{2}$Aurora Innovation Inc. \texttt{\small sanjiban.choudhury@gmail.com}}%
\thanks{This work was (partially) funded by the National Science Foundation IIS (\#2007011), National Science Foundation DMS (\#1839371), the Office of Naval Research, US Army Research Laboratory CCDC, Amazon, and Honda Research Institute USA.}
}

\begin{document}

\maketitle
\thispagestyle{empty}
\pagestyle{empty}

\begin{abstract}
    
    Sampling-based motion planners rely on incremental densification to discover progressively shorter paths. 
    After computing feasible path $\xi$ between start $x_s$ and goal $x_t$,
    the \emph{Informed Set} (IS) prunes the configuration space $\Cspace$ 
    by conservatively eliminating points that cannot yield shorter paths.
    Densification via sampling from this Informed Set retains asymptotic optimality of sampling from the entire configuration space.
    For path length $c(\xi)$ and Euclidean heuristic $h$,
    $IS = \{ x | x \in \Cspace, h(x_s, x) + h(x, x_t) \leq c(\xi) \}$.
    
    Relying on the heuristic can render the IS
    especially conservative in high dimensions or complex environments.
    Furthermore, the IS only shrinks when shorter paths are discovered.
    Thus, the computational effort from each iteration of densification and planning is wasted if it fails to yield a shorter path,
    despite improving the cost-to-come for vertices in the search tree.
    Our key insight is that even in such a failure,
    shorter paths to \emph{vertices in the search tree} (rather than just the goal) can immediately improve the planner's sampling strategy.
    Guided Incremental Local Densification (\guild{}) leverages this information to sample from \emph{Local Subsets} of the IS.
    We show that \guild{} significantly outperforms uniform sampling of the Informed Set in simulated $\R^2$, $SE(2)$ environments and manipulation tasks in $\R^7$.
\end{abstract}
\section{Introduction}
\label{sec:intro}

Sampling-based algorithms have shown tremendous success in 
solving complex high-dimensional robot motion planning problems.
These algorithms achieve asymptotic-optimality by incrementally densifying the robot's configuration space to continually improve the shortest path in an anytime manner~\cite{KF11,karaman10Incremental,ArslanT13,GSB15,StrubG20ABIT,StrubG20AIT}.
While such algorithms often compute an initial feasible path quickly, their convergence is slow because they need to sample a huge number of configurations before a shorter path is found. 

\begin{figure}[!t]
\centering
\includegraphics[width=\linewidth]{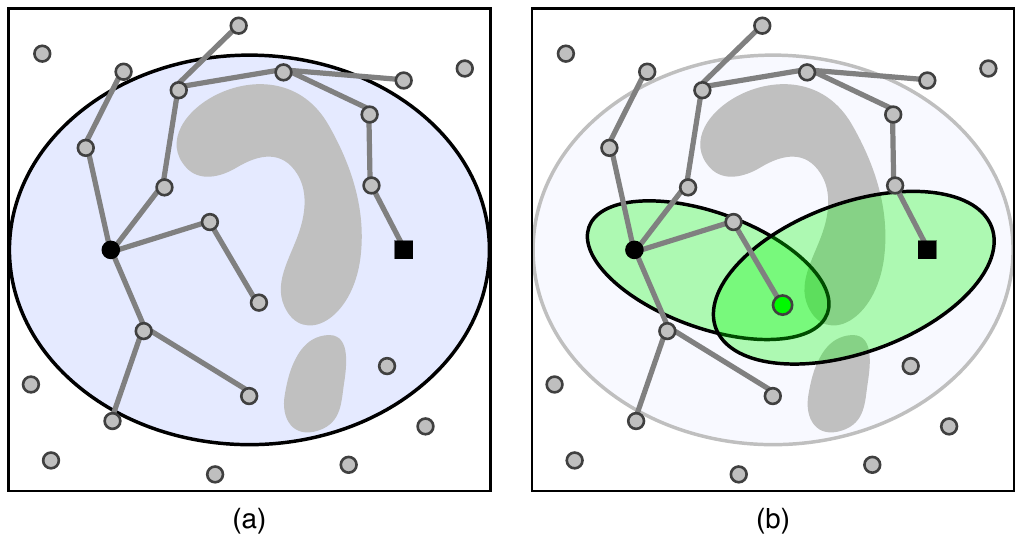}
\caption{
Comparison of (a) Informed Set and (b) \emph{Local Subsets} induced by \guild{}.
To further focus sampling within the Informed Set,
\guild{} chooses a beacon (green) and
decomposes the original problem into two smaller problems
using information from the search tree.
}
\label{fig:guild_intro}
\end{figure}

While the configuration space can be densified uniformly to find shorter paths, 
the current best path $\Path{}$ between the start $\xs$ and target $\xg$ defines
an \emph{Informed Set} $\Cinf{} \subseteq \Cspace{}$~\cite{gammell2014informed} that 
contains only states that can yield shorter paths:
\begin{equation}
\label{eq:informedset}
\Cinf{} = \{x \mid x \in \Cspace{}, h(\xs, x) + h(x, \xg) \leq c(\Path)\},
\end{equation}
where $c(\Path)$ is the cost of the current path and $h$ is an admissible heuristic. 
For problems minimizing path length in $\R^n$,
the Informed Set with the Euclidean distance heuristic is 
an $n$-dimensional prolate hyperspheroid $\ellipseinf{}$,
parameterized by foci $\xs$  and $\xg$ and transverse axis diameter $c(\Path)$ 
(i.e., a generalized ellipse as seen in \Cref{fig:guild_intro}a).
Sampling from $\ellipseinf{}$ preserves the asymptotic optimality guarantee of sampling from the entire configuration space $\Cspace{}$.

In practice, however,
the Informed Set often provides limited sample efficiency improvement.
Since the measure of the Informed Set $\lambda(\ellipseinf{})$ is a function of the current solution cost, 
a high cost solution may yield a large Informed Set with comparable (or greater) measure to the full state space $\lambda(\Cspace{})$.
Paradoxically,
to reap the greatest benefits of sampling from the Informed Set,
the planner must already 
have a path of sufficiently small cost.
This is a particular challenge for planning problems in high dimensions or cluttered environments.
Furthermore, 
the computational effort from each iteration
of densification and planning is wasted if it fails to yield a
shorter path.

\begin{figure*}[!ht]
\centering
\includegraphics[width=\linewidth]{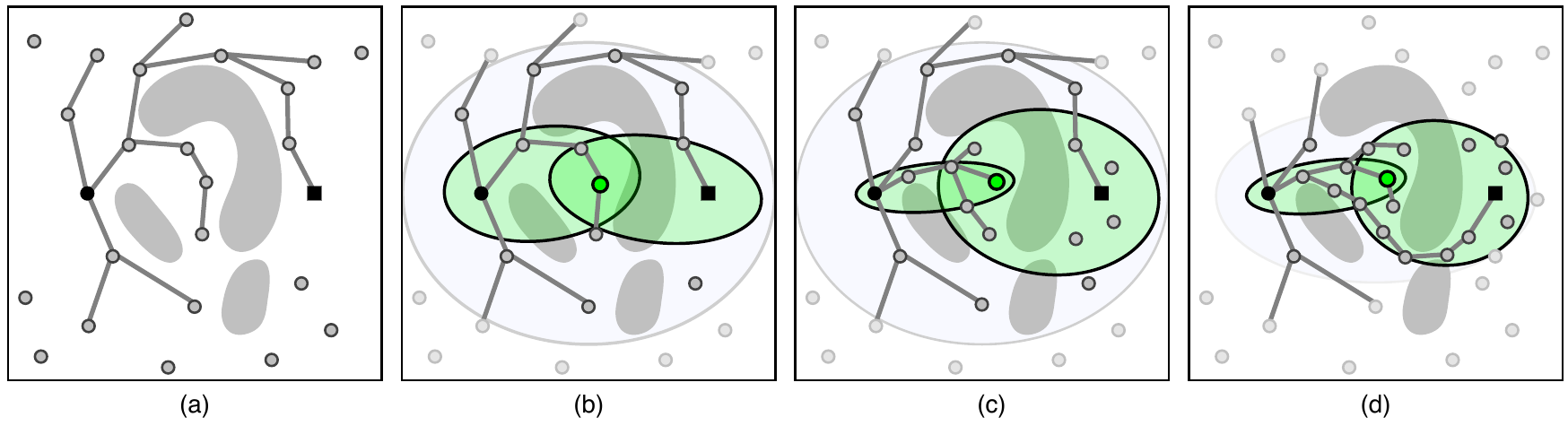}
\caption{
\textbf{(a)}
Initial solution.
\textbf{(b)}
\guild{} selects a beacon and induces Local Subsets (green), which do not cover the narrow passage.
\textbf{(c)}
The planner does not find a shorter path to the goal, so the Informed Set is unchanged.
However, \guild{} leverages the improved cost-to-come in the search tree to update the Local Subsets.
The start-beacon set shrinks to further focus sampling,
and the remaining slack between the beacon's and goal's cost-to-comes is used to \emph{expand} the beacon-target set.
The Local Subsets now cover the narrow passage,
focusing sampling within the Informed Set to quickly converge to
\textbf{(d)}
the optimal solution.
}
\label{fig:ideal_scenario}
\vspace{2\baselineskip}
\end{figure*}

Our key insight is that even in such a failure,
we can open the black box of the planning algorithm 
to immediately improve the planner's sampling strategy.
With
\textbf{Gu}ided \textbf{I}ncremental \textbf{L}ocal \textbf{D}ensification (\guild{}),
shorter paths to \emph{any vertex in the search tree} can guide further sampling.
\guild{} 
introduces the idea of a beacon, 
a vertex in the search tree that
decomposes the original sampling/planning problem into two smaller subproblems (\Cref{fig:guild_intro}b).
Much like the Informed Set between the start and target,
the beacon induces \emph{Local Subsets} with
(i) start and beacon as foci and 
(ii) beacon and target as foci. 
\guild{} leverages improvements to the search tree to adapt the Local Subsets
and converge to the optimal path with fewer samples
(\Cref{fig:ideal_scenario}).

\newpage

We make the following contributions:
\begin{itemize}
\item We introduce \guild{},
an incremental densification framework that
effectively leverages search tree information to
focus sampling.
\item We compare theoretical properties of
the Local Subsets that \guild{} samples to
the Informed Set.
\item We propose several \BeaconSelector{} strategies for \guild{},
including an adversarial bandit algorithm.
\item We show experimentally that regardless of the \BeaconSelector{},
\guild{} outperforms the state-of-the-art Informed Set densification baseline
across a range of planning domains.
In particular, \guild{} 
yields modest improvements in simpler planning domains and
excels in domains with difficult-to-sample homotopy classes.
\end{itemize}

\section{Related Works}

Sampling-based algorithms construct roadmaps~\cite{kavraki96prm} or 
trees~\cite{lavallek01,kuffner2000rrt,hsuEST} by sampling configurations 
and connecting the nearest neighbors to compute a collision-free shortest path between the start and goal. 
Sampling-based algorithms for \emph{optimal} motion planning incrementally densify the 
configuration space to determine shorter paths in an anytime manner~\cite{KF11,JSCP15,GSB15,StrubG20ABIT,StrubG20AIT,ArslanT13} 
guaranteeing asymptotic optimality.
However, their convergence is typically slow,
especially in higher dimensions and complex environment with narrow passages 
where a uniform sampler requires $O((\frac{1}{\delta})^d)$ configurations 
to compute the optimal path with $\delta$-clearance in a $d$-dimensional space, 
which can be computationally prohibitive~\cite{hsuEST}.

This motivated literature that focused on improving the sampling strategy.
A popular approach is to extract information from the workspace to 
focus sampling in bottleneck regions and narrow passages.
While some algorithms in this class sample \emph{between} regions of collision~\cite{wilmarth1999maprm, holleman2000framework, hsu2003bridge, kurniawati2004workspace, yang2004adapting, van2005using,DennyGTA14} 
to identify narrow passages, 
others sample close to obstacles to compute 
high quality paths expected to follow the contours of the obstacles~\cite{amato1998obprm,boor1999gaussian,amatoOBRRT}.
While these samplers reduce the number of configurations required to compute a solution, they perform computationally-intensive geometric tests to accept samples into the roadmap, resulting in large planning times.
Nevertheless, these approaches can be used in parallel to \guild{} to sample non-uniformly within 
the Local Subsets and further improve convergence.

Recently, to alleviate the issues with approaches relying on explicit geometric tests, 
there has been effort in determining low-dimensional structure in the workspace to bias sampling.
Generative modeling tools~\cite{doersch2016tutorial} have been applied 
to learn sampling distributions, conditioned upon 
the current planning problem and the obstacle distribution~\cite{ichter2017learning,kumar2019lego,IchterSLF20,chamzas2019primitives,chamzas3D,arbaazGNN}.
These approaches do not explicitly reason about the Informed Set to provide optimality guarantees, 
and fail to robustly sample bottleneck regions in complex environments.
However, they share with \guild{} the underlying idea of identifying critical samples
and can be leveraged to inform beacon selection.

A different class of algorithms adapt sampling online as planning progresses.
Toggle-PRM~\cite{DennyA12} constructs one roadmap in the free space and another in the obstacle space to infer 
narrow passages and increase sampling density in such regions.
Some approaches trade-off between exploration of the configuration space
and exploitation of the search tree~\cite{diankov2007randomized} or the underlying cost space~\cite{JailletCS10} to bias sampling online.
DRRT~\cite{HauerT17} guides sampling by following a gradient of the underlying cost function.
Guided-EST~\cite{PhillipsBK04} samples new configurations by considering cost as well as the current local sampling density.
Relevant Regions~\cite{ArslanT15RR,ArslanT15ML,JoshiT20} improve the performance of EST~\cite{hsuEST} by using information from the search tree to restrict sampling within the Informed Set.
\guild{} best fits in this class of algorithms that bias sampling with the state of the planning algorithm.

\section{Sampling-Based Optimal Motion Planning via Incremental Densification}
\label{sec:problem_formulation}

In this section, we formally introduce the problem of sampling-based optimal motion planning (optimal SBMP).
Let $\calX \subseteq \R^n$ be the statespace of the planning problem,
where $\Cobs \subset \calX$ is the subspace occupied by obstacles and
the free space is $\Cfree = \calX \backslash \Cobs$.
Given source and target states $\xs, \xg \in \Cfree$,
a path $\Path: [0,1] \rightarrow \Cspace{}$ is represented as a sequence of states such that $\Path(0) = \xs$ and $\Path(1) = \xg$.
Let $\PathSet{}(\xs, \xg)$ be the set of all such paths.
Given a cost function $c: \Path \rightarrow \R_{> 0}$, an optimal collision-free path is defined as:
\begin{align*}
\Path^* &= \argmin_{\Path \in \PathSet{}(\xs, \xg)}~c(\Path)
           \hspace{3mm} s.t. \hspace{3mm}
           \forall t \in [0,1],~\Path(t) \in \Cfree{}.
\end{align*}

Optimal SBMP algorithms
progressively improve a roadmap approximation of the state space
to plan asymptotically-optimal paths (\Cref{alg:planner}).
In each iteration, states are sampled from \Cfree{} to grow the vertices of an edge-implicit\footnotemark{} graph $\graph{}$ (Line 4).
\footnotetext{
The radius of connectivity of the implicit graph is chosen to ensure asymptotic optimality~\cite{KF11}. 
}
Then, a shortest path algorithm computes the \emph{resolution-optimal} path on $\graph{}$,
internally using an admissible heuristic $h$ to focus the search (Line 5).
If densification produces a lower-cost path, the best solution cost is updated and that path is emitted as $\Path_i$ (Lines 6-8).
As the discrete graph $\graph{}$ more closely approximates the continuous state space,
the sequence of resolution-optimal paths $\{\Path_1, \Path_2, \cdots \}$ approaches the optimal path $\Path^*$.

\subsection{Informed Incremental Densification}

Given a candidate state $x \in \Cfree$ and the admissible search heuristic $h$,
the cost of all paths between $\xs$ and $\xg$ that pass through $x$
is lower bounded by $h(\xs, x) + h(x, \xg)$. 
Defined by \Cref{eq:informedset},
the Informed Set (IS)
excludes states for which this lower bound exceeds the best solution cost $c(\Path)$.
Unlike sampling the entire state space $\Cfree{}$,
sampling from the IS automatically excludes states that cannot be part of a lower-cost path \cite{gammell2014informed}.

Euclidean distance, an admissible heuristic for path length, 
admits a concise geometric interpretation of the IS:
an $n$-dimensional prolate hyperspheroid
$\ellipseinf{} = \ellipse(\xs, \xg, c(\Path))$ with
foci $\xs, \xg$ and transverse axis diameter $c(\Path)$.
\begin{equation}
\label{eq:l2informedset}
\ellipseinf{} = \{x \mid x \in \Cspace{}, \|\xs - x\|_2 + \|x - \xg\|_2 \leq c(\Path) \}
\end{equation}
Prolate hyperspheroids can be sampled analytically, rather than via rejection sampling~\cite{gammell2014informed}.

Sampling from the IS is a sufficient condition to converge to $\Path^*$.
The improved efficiency can be characterized by comparing
the measure $\lambda(\ellipseinf{})$ to the measure of the full state space $\lambda(\Cspace{})$.
When $\lambda(\ellipseinf{}) < \lambda(\Cspace)$,
the IS can yield significant improvements on sample efficiency.
However, when the initial solution has high path length (e.g., due to a cluttered environment),
$\lambda(\ellipseinf)$ may instead be closer to---or even larger than---$\lambda(\Cspace)$.
This issue is exacerbated because the IS only shrinks when shorter paths are found.
As a result,
each iteration of densification and search that fails to find a shorter path does not affect the state sampling distribution.
In the next section, we introduce a new strategy that leverages this previously-wasted computational effort.


%

\begin{algorithm}[t]
\SetAlgoLined
\setstretch{1.2}
\caption{Informed Optimal SBMP}
\label{alg:planner}
\SetKwInOut{Input}{Input}
\SetKwInOut{IOutput}{IOutput}
\SetKwInOut{Output}{Output}
\Input{start~$\vs$, goal~$\vg$}
\Output{$\{\Path_1,~\Path_2,\cdots\}$ s.t. $c(\Path_{i+1}) < c(\Path_i)$}
Initialize best solution cost: $c(\Path) \gets \infty$ \\ 
Initialize edge-implicit graph: $\graph{} \gets \{\vs,\vg\}$ \\
\Repeat{forever}
{
    \textcolor{red}{Densify: $\graph{} \xleftarrow{+} $ \texttt{Sample}$(\vs,~\vg,~c(\Path))$}  \\
    Compute shortest path: $\hat{\Path} \leftarrow$ \texttt{Search}$(\vs,~\vg,~\graph)$\\
    \If{$c(\hat{\Path}) = g(\vg) < c(\Path)$}
    {
        Update best solution cost: $c(\Path) \gets g(\vg)$ \\
        Emit current solution $\hat{\Path}$ \\ 
    }
}
\end{algorithm}

\section{Guided Incremental Local Densification}
\label{sec:approach}


We present \guild{},
an incremental densification framework
that leverages partial search information
to focus sampling within the IS.
If an iteration of densification and planning has failed to improve the solution cost,
what information is there for \guild{} to take advantage of?

Our key insight is to open the black box of the underlying search algorithm.
Although the iteration may not have found a shorter path to the goal,
new shorter paths to \emph{other vertices in the search tree}
can immediately improve the sampling strategy (\Cref{alg:planner}, Line 4).

\subsection{Guiding Densification with Search Tree Information}
\label{sec:guild-tree}

During search (\Cref{alg:planner}, Line 5), the algorithm internally expands vertices in \graph{} to construct a search tree $\searchTree{}$.
Each vertex $v \in \searchTree$ is associated with
a parent vertex,
a cost-to-come $g(v)$ via that parent, and
a cost-to-go heuristic estimate $h(v, \vg)$.
When new states are sampled and added to $\graph$,
they may also be added to $\searchTree$.
Other vertices may then discover
that their cost-to-come would be reduced
by updating their parent to this new vertex~\cite{KF11,GSB15}.
While the IS only shrinks when these changes propagate all the way to $\vg$,
\guild{} leverages \emph{any} cost-to-come improvement to adaptively guide densification.

\begin{figure}[!t]
\centering
\includegraphics[width=0.6\linewidth]{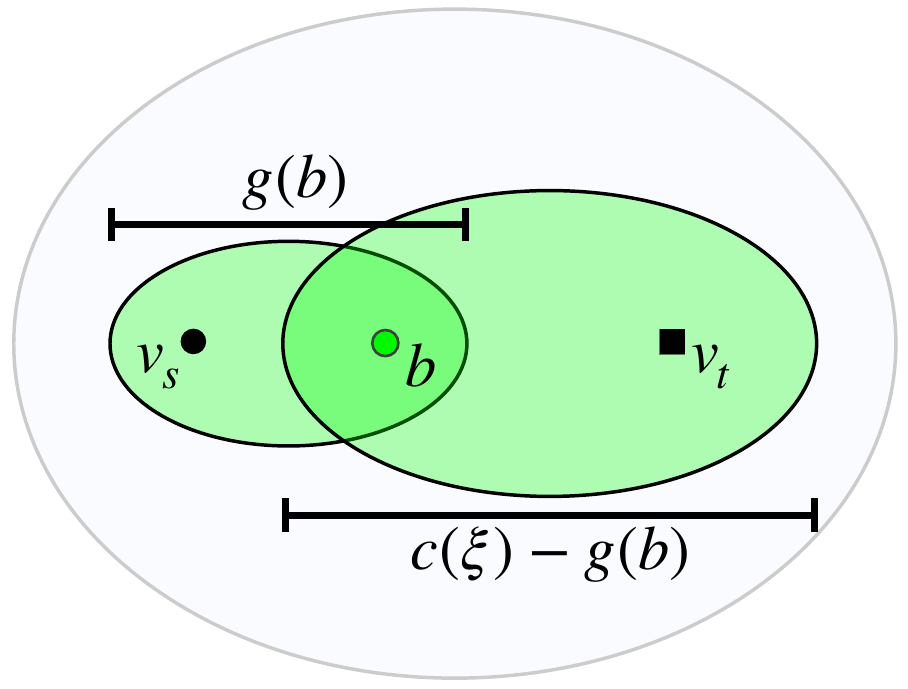}
\caption{
Local Subsets $\ellipselocal{}$ (green) are defined by a beacon $b$,
as well as the cost-to-come on the search tree $g(b)$
and the current best solution cost $c(\Path)$.
}
\label{fig:LS_Construction}
\vspace{1\baselineskip}
\end{figure}

\guild{} 
introduces the idea of a \emph{beacon}:
a vertex in the search tree that
decomposes the original sampling/planning problem into two smaller subproblems
(\Cref{fig:LS_Construction}).
We define the \emph{Local Subsets} (LS)
induced by beacon $b$ to be the union of two prolate hyperspheroids,
with foci $(\vs, b)$ and foci $(b, \vg)$.
The start-beacon set has transverse axis diameter $g(b)$,
only including points that can improve the cost-to-come from $\vs$ to $b$.
Given the current shortest subpath to $b$ and its cost-to-come,
the beacon-target set only includes points that can extend that subpath and improve the solution cost
by setting the transverse axis diameter to $c(\Path) - g(b)$.

\guild adapts $\ellipselocal{}$ after each iteration of densification and planning.
When $g(b)$ is reduced, the start-beacon set shrinks and the beacon-target set \emph{expands}.
Surprisingly, this expansion is actually desirable:
if the best solution cost remains the same and the beacon's cost-to-come is reduced,
there is a larger path length budget that can be expended between the beacon and target
that could still yield a shorter path overall (\Cref{fig:ideal_scenario}).

We summarize the \guild framework in \Cref{alg:guild},
which replaces the highlighted line in \Cref{alg:planner}.
The \BeaconSelector (\Cref{alg:guild}, Line 4)
is a function that chooses a beacon to guide local densification.
The beacon must have been expanded by the underlying search algorithm to preserve the sampling guarantees that we prove in the next section.\footnotemark{}
\guild samples uniformly from $\ellipselocal{}$ (Line 6)
using the same analytic strategy proposed for $\ellipseinf{}$ \cite{GSB20}.
\footnotetext{
\guild must sample from the IS with nonzero probability to retain asymptotic optimality.
This is achieved by including $\vs$ in the beacon set.
}

In \Cref{alg:selector_definitions},
we present some candidate beacon selectors
that we evaluate in \Cref{sec:experiments}.
The \texttt{InformedSet} beacon selector recovers the behavior of sampling from the IS by choosing $\vs$ as the beacon.
\uniformguild uniformly samples from the beacon set $\beaconSet$.
The \greedyguild beacon selector aims to select the beacon
with the maximum possible improvement in path length,
while minimizing the measure of the set that needs to be sampled.
It takes the ratio of these two quantities:
\begin{align}
\label{eq:greedy-objective}
w(b) &= \frac{c(\Path) - h(\vs, b) - h(b, \vg)}{\lambda(\ellipselocal{})}.
\end{align}
Finally, the \banditguild beacon selector implements the EXP3 adversarial bandit algorithm \cite{auer2002}.
The bandit reward function is the fractional improvement in the path length
\begin{align*}
r(b) &= \frac{c(\Path_{i-1}) - c(\Path_i)}{c(\Path_{i-1})}.
\end{align*}
An adversarial bandit algorithm is necessary because this reward function is nonstationary.

\begin{figure*}[!ht]
\setlength{\fboxrule}{1pt}
\setlength{\fboxsep}{0pt}
\newlength{\envwidth}
\setlength{\envwidth}{0.18\linewidth}
\centering
\begin{subfigure}{\envwidth}
\fbox{\includegraphics[height=\linewidth]{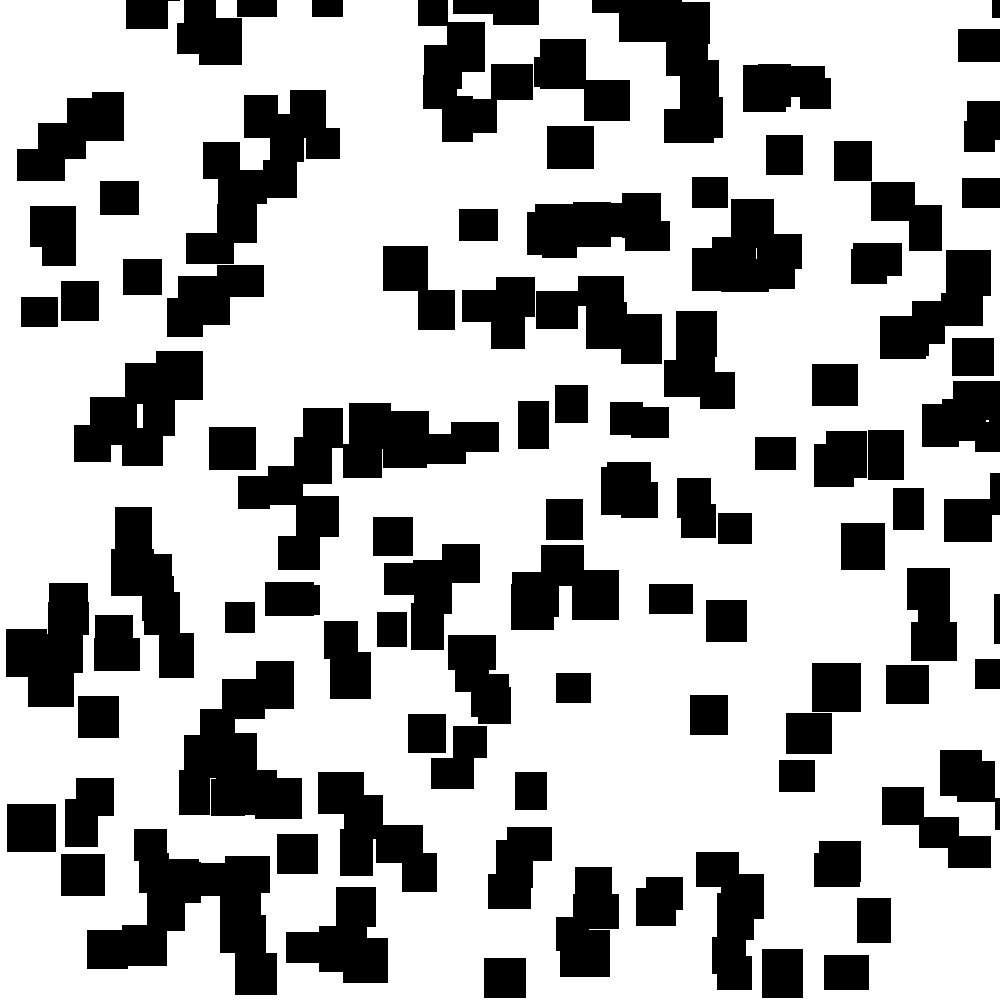}}
  \caption{\forestenv}
  \label{fig:forestenv}
\end{subfigure}
\begin{subfigure}{\envwidth}
\fbox{\includegraphics[height=\linewidth]{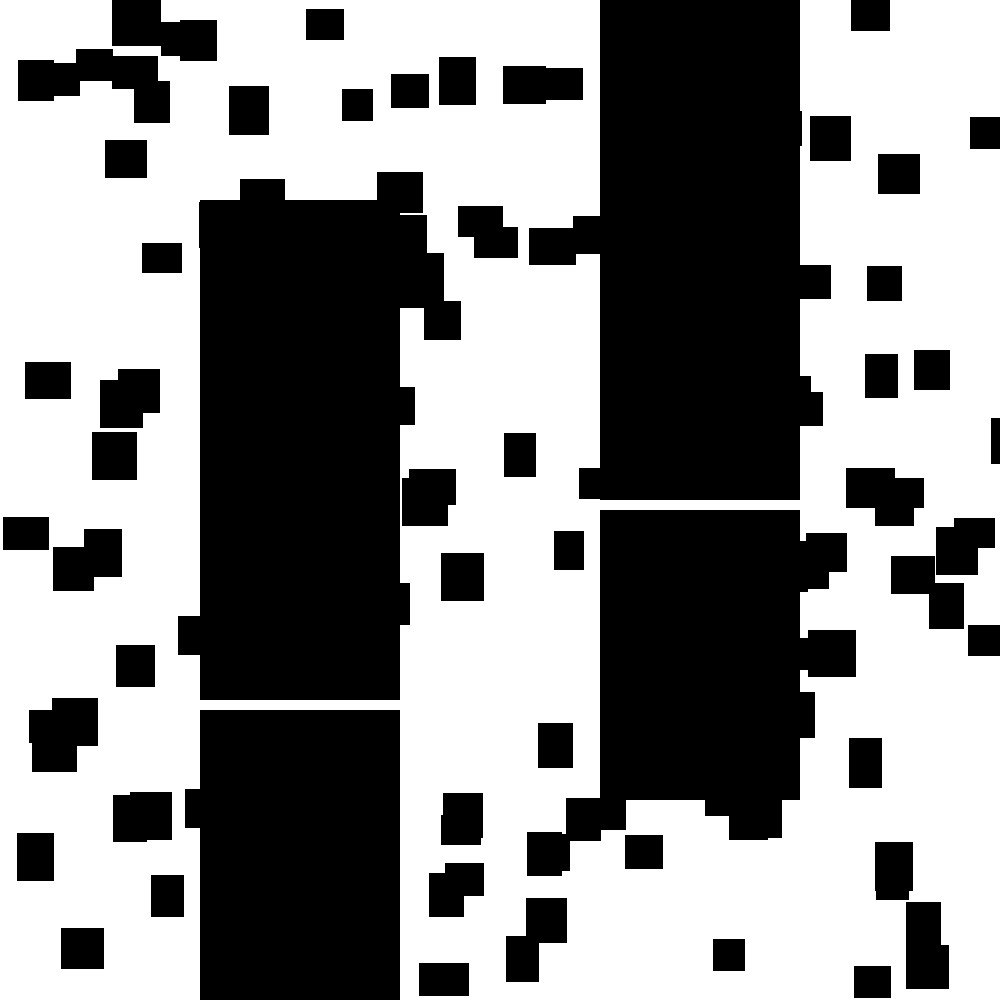}}
  \caption{\twowallforestenv}
  \label{fig:twowallforestenv}
\end{subfigure}
\begin{subfigure}{\envwidth}
\fbox{\includegraphics[height=\linewidth]{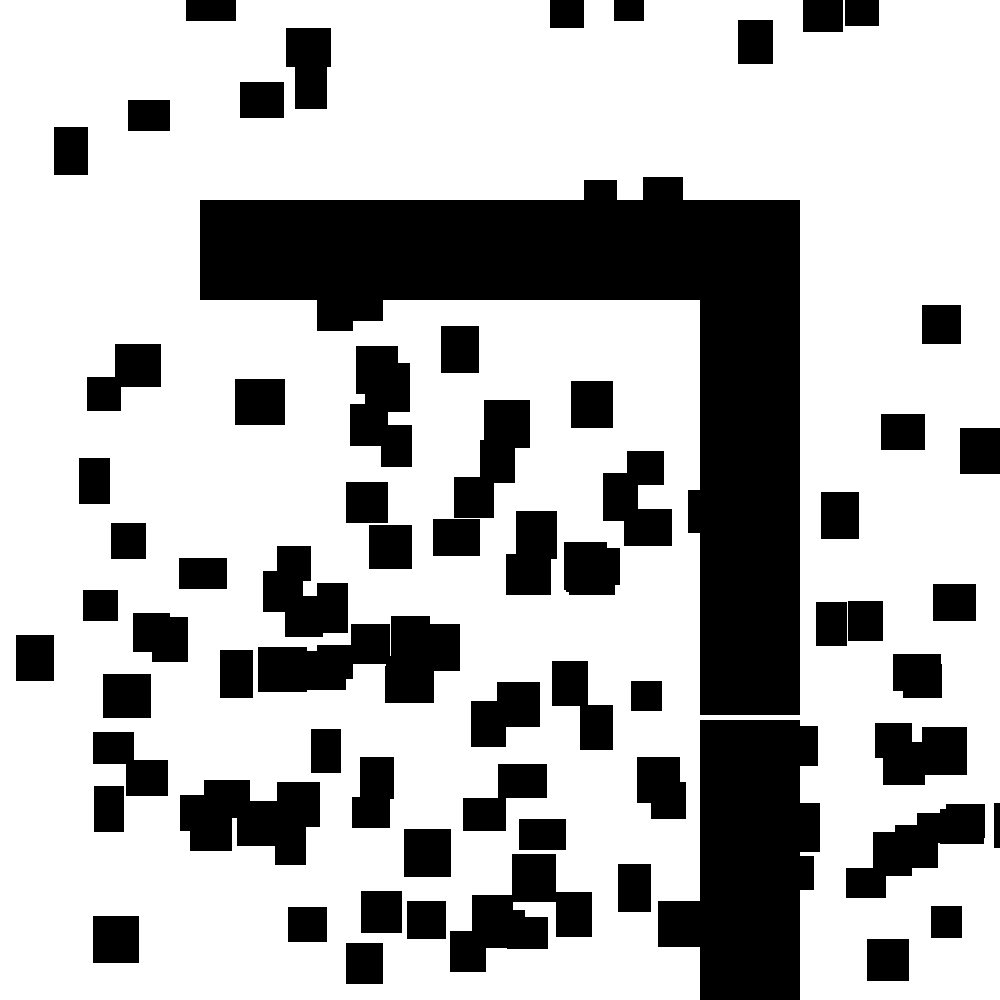}}
  \caption{\lforestenv}
  \label{fig:lforestenv}
\end{subfigure}\hspace{2em}
\begin{subfigure}{\envwidth}
\fbox{\includegraphics[height=\linewidth]{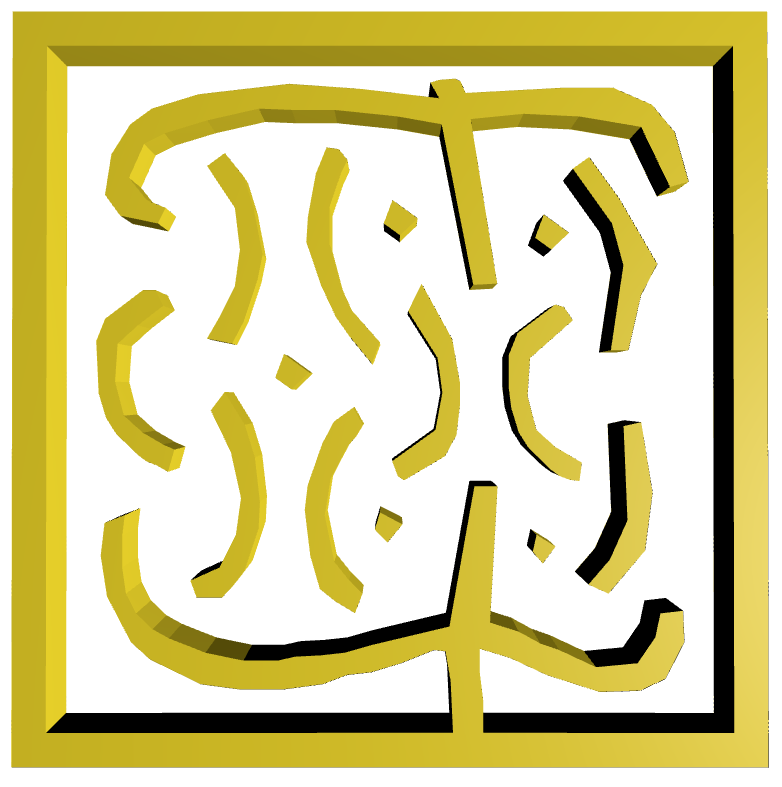}}
  \caption{\carenv}
  \label{fig:carenv}
\end{subfigure}
\begin{subfigure}{\envwidth}
\fbox{\includegraphics[height=\linewidth]{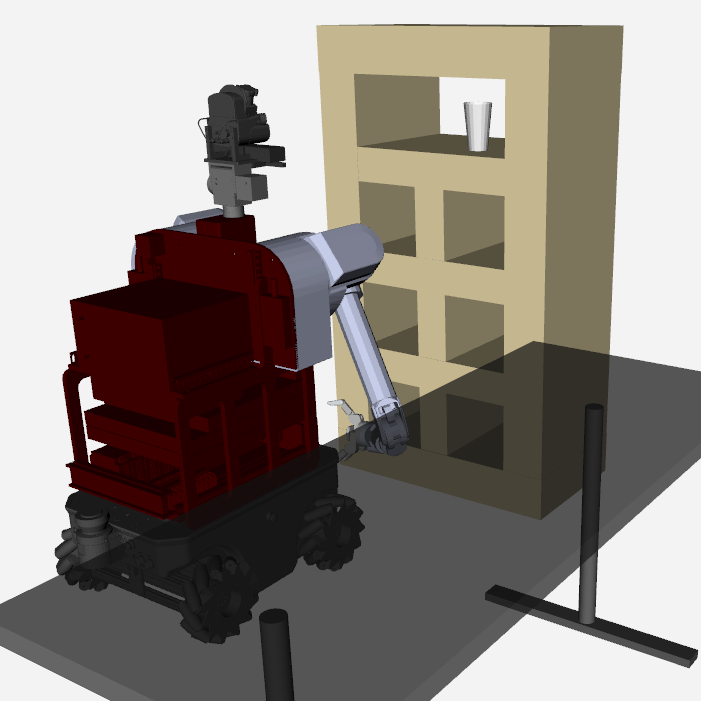}}
  \caption{\herbenv}
  \label{fig:herbenv}
\end{subfigure}
\caption{
Evaluation environments.
}
\label{fig:envs}
\setlength{\fboxrule}{0.2pt}
\setlength{\fboxsep}{3pt}
\vspace{-1\baselineskip}
\end{figure*}

\subsection{Properties of Local Subsets}
\label{sec:local-subsets-theory}

First, we guarantee that
\guild will never draw samples outside the IS by
showing that both sets in $\ellipselocal{}$ are subsets of the IS (\Cref{theorem:subset}).
Then, we show that the measure of $\ellipselocal{}$ is upper-bounded by that of the IS (\Cref{thm:union:subset}),
so choosing the appropriate beacon will allow \guild to sample the space more efficiently.


\begin{theorem}
\label{theorem:subset}
Given the current best solution cost $c(\Path_i)$,
a beacon $b \in \searchTree$ previously expanded with cost-to-come $g(b)$, 
we have that $\ellipse(\vs, b, g(b)),~\ellipse(b, \vg, c(\Path_i)-g(b)) \subseteq \ellipseinf$.
\end{theorem}

\begin{proof}
Let $x \in \ellipse(\vs, b, g(b))$.
By construction of the prolate hyperspheroid and the triangle inequality:
\begin{align*}
    h(\vs, x) + h(x, b) &\leq g(b) \\
    h(\vs, x) + h(x, b) + h(b, \vg) &\leq g(b) + h(b, \vg) \\
    h(\vs, x) + h(x, \vg) &\leq g(b) + h(b, \vg)
\end{align*}
Since $b$ was previously expanded, 
\begin{align*}
    h(\vs, x) + h(x, \vg) &\leq g(b) + h(b, \vg) \leq c(\xi_i)
\end{align*}
Therefore, we have $\ellipse(\vs, b, g(b)) \subseteq \ellipseinf{}$.

Similarly, let $x \in \ellipse(b, \vg, c(\xi_i) - g(b))$. We have,
\begin{align*}
    h(b, x) + h(x, \vg) &\leq c(\xi_i) - g(b) \\
    g(b) + h(b, x) + h(x, \vg) &\leq c(\xi_i) \\
    h(\vs, x) + h(x, \vg) &\leq c(\xi_i)
\end{align*}
Therefore, we have $\ellipse(b, \vg, c(\xi_i) - g(b)) \subseteq \ellipseinf{}$.
\end{proof}

\begin{algorithm}[!t]
\SetAlgoLined
\setstretch{1.2}
\caption{\guild{}}
\label{alg:guild}
\SetKwInOut{Input}{Input}
\SetKwInOut{IOutput}{IOutput}
\SetKwInOut{Output}{Output}
\Input{beacon set $\beaconSet{}$, search tree $\searchTree$, graph $\graph{}$}
\Output{densified graph $\graph{}$}
\If{a solution does not yet exist}
{
    $\graph{} \xleftarrow{+}$ \texttt{UniformSample(\Cfree)}
}
\Else
{
    $b \gets \BeaconSelector(\beaconSet)$ \\
    $\ellipselocal{} \gets \ellipse(\vs, b, g(b))~\cup~ \ellipse(b, \vg, c(\Path) - g(b))$ \\
    $\graph{} \xleftarrow{+} \texttt{UniformSample}(\ellipselocal{})$ 
}
\Return $\sampleSet$
\end{algorithm}
\begin{algorithm}[!t]
\setstretch{1.2}
\SetAlgoLined
\caption{Candidate \BeaconSelector{}s}
\label{alg:selector_definitions}

\SetKwInOut{Input}{Input}
\SetKwInOut{IOutput}{IOutput}
\SetKwInOut{Output}{Output}
\SetKwProg{function}{Function}{}{}

\SetKwFunction{InformedSet}{InformedSet}
\SetKwFunction{Uniform}{Uniform}
\SetKwFunction{Greedy}{Greedy}
\SetKwFunction{Bandit}{Bandit}

\Input{beacon set $\beaconSet{}$}
\vspace{2mm}

\function{\InformedSet}
{
	\KwRet\ \vs;
}
\vspace{1mm}
\function{\Uniform}
{
	\KwRet\ $b \sim \calU(\beaconSet{})$;
}
\vspace{1mm}
\function{\Greedy}
{
  \KwRet\ $\argmax_{b \in \beaconSet{}}{w(b)}$
  \Comment{\Cref{eq:greedy-objective}}
}
\vspace{1mm}
\function{\Bandit}
{
  \KwRet\ EXP3$(\beaconSet{})$
}
\end{algorithm}


\begin{theorem}
\label{thm:union:subset}
Given the current best solution cost $c(\xi_i)$, a beacon $b \in \searchTree$ previously expanded with cost-to-come $g(b)$, we have $\lambda(\ellipselocal) \leq \lambda(\ellipseinf)$
\end{theorem}

\begin{proof}
A prolate hyperspheroid $\ellipse \in \R^n$
with transverse diameter $a$ and
distance between the foci $f$
has measure
$\lambda(\ellipse) = \calK a \left(a^2-f^2\right)^{\frac{n-1}{2}}$
where $\calK = \frac{\pi^{\frac{n}{2}}}{2^n\Gamma(\frac{n}{2} + 1)}$ is only dimension-dependent.
Let $a_s,f_s$ and $a_t,f_t$ denote
the parameters corresponding to
$\ellipse(\vs, b, g(b))$ and $\ellipse(b, \vg, c(\Path_i) - g(b))$ respectively.
We have the following:
\begin{align*}
    a_s + a_t &= a_{\rm IS} \hspace{5mm} \text{(by definition)}\\
    f_s + f_t &\geq f_{\rm IS} \hspace{5mm} \text{(triangle inequality)}
\end{align*}
Consider the measure of the IS:
\begin{align*}
    \lambda(\ellipseinf) &= \calK a_{\rm IS} \left(a_{\rm IS}^2-f_{\rm IS}^2\right)^{\frac{n-1}{2}} \\
    &\geq \calK (a_s+a_t)\left((a_s + a_t)^2-(f_s+f_t)^2\right)^{\frac{n-1}{2}} \\
    &\geq \calK (a_s+a_t)\left(a_s^2 + a_t^2-f_s^2-f_t^2\right)^{\frac{n-1}{2}} \\
    &\geq \calK (a_s+a_t)\left((a_s^2 - f_s^2) + (a_t^2 - f_t^2)\right)^{\frac{n-1}{2}} \\
    &\geq \calK a_s(a_s^2-f_s^2)^{\frac{n-1}{2}} + \calK a_t(a_t^2-f_t^2)^{\frac{n-1}{2}}
\end{align*}
Therefore, we have $\lambda(\ellipseinf) \geq \lambda(\ellipselocal)$.
\end{proof}

\section{Experiments}
\label{sec:experiments}

We evaluate \guild{} on an array of planning problems to characterize the proposed beacon selectors (\Cref{fig:envs}).

In the $\R^2$ environments, the task is to plan from the bottom left corner
to the top right corner (\forestenv, \twowallforestenv),
or the bottom right corner (\lforestenv).
A forest of obstacles is randomly placed throughout each environment.
The easier \forestenv environment only has these random obstacles;
as a result, there are many different homotopy classes that will produce near-optimal paths.
\twowallforestenv and \lforestenv introduce large obstacles with narrow passages that must be crossed to produce near-optimal paths.
Discovering these passages typically requires the space to be sampled very densely.
At the lower sampling resolutions resulting from fewer initial graph samples, there are many suboptimal homotopy classes that are more easily sampled and discovered.
Therefore, initial paths will have high cost and $\lambda(\ellipseinf)$ will be much larger than $\lambda(\Cspace)$.
These more challenging scenarios highlight the limitations of IS densification.

In \carenv, a benchmark environment from OMPL~\cite{ompl}, the task is to navigate through a maze.
In \herbenv, a 7-DOF manipulator \cite{srinivasa2010herb} is tasked with moving its end-effector to pick an object from a shelf.

\subsection{Evaluation Metrics and Hypotheses}

The first metric we consider is the \emph{Sample Efficiency} of incremental densification:
how many samples must be drawn before the optimal SBMP algorithm converges to the cost of the optimal path?
We determine this minimum cost $c(\Path^*)$ by running with a large timeout
(1 min. for $\R^2$ problems, 5 min. for higher-dimensional planning problems).

\begin{hypothesis}
\item \label{hyp:sample-efficiency}
  \emph{\guild will require fewer samples to converge to the minimum solution cost than the Informed Set.}
\end{hypothesis}

Next, to understand the performance of the optimal SBMP algorithm over time,
we then consider the \emph{Convergence Percentage} and \emph{Normalized Path Cost} as a function of samples.
Convergence Percentage is the fraction of trials where the planner had converged to the optimal path cost, with that number of samples.
Normalized Path Cost divides the cost of the current best solution by the optimal path cost $\frac{c(\Path)}{c(\Path^*)}$.

\begin{hypothesis}
\item \label{hyp:success-ratio}
  \emph{For a fixed sample budget, \guild will have a higher Convergence Percentage than the Informed Set.}
\item \label{hyp:path-cost}
  \emph{For a fixed sample budget, \guild will have lower Normalized Path Cost than the Informed Set.}
\end{hypothesis}

\begin{figure}[!t]
\centering
\setlength{\fboxrule}{1pt}
\setlength{\fboxsep}{0pt}
\newlength{\heatmapwidth}
\setlength{\heatmapwidth}{0.22\linewidth}
\fbox{\includegraphics[width=\heatmapwidth]{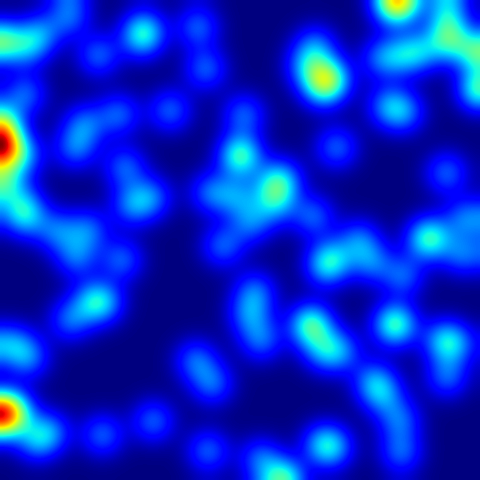}}
\fbox{\includegraphics[width=\heatmapwidth]{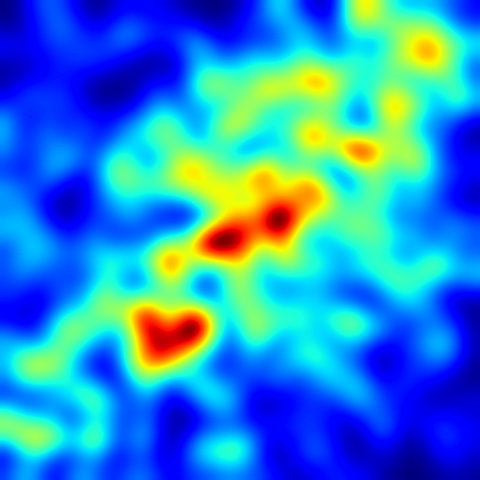}}
\fbox{\includegraphics[width=\heatmapwidth]{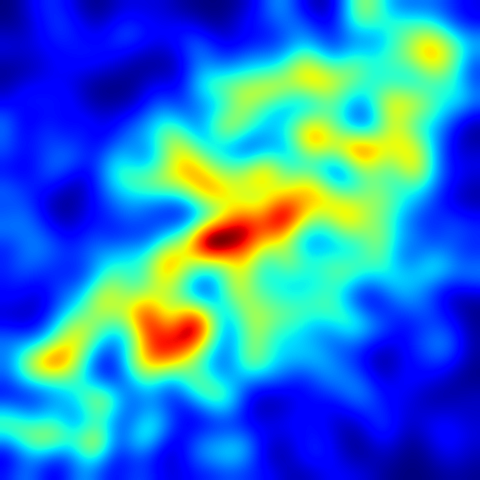}}
\fbox{\includegraphics[width=\heatmapwidth]{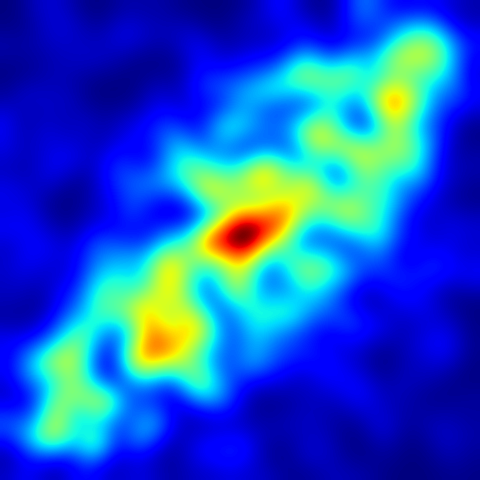}}

\medskip

\fbox{\includegraphics[width=\heatmapwidth]{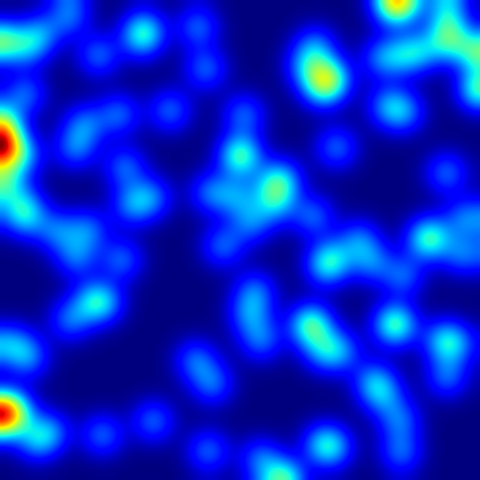}}
\fbox{\includegraphics[width=\heatmapwidth]{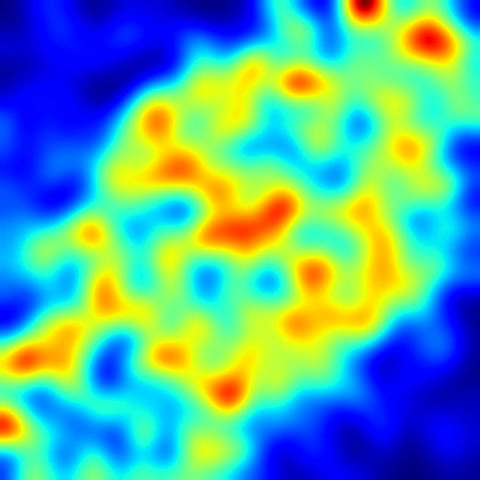}}
\fbox{\includegraphics[width=\heatmapwidth]{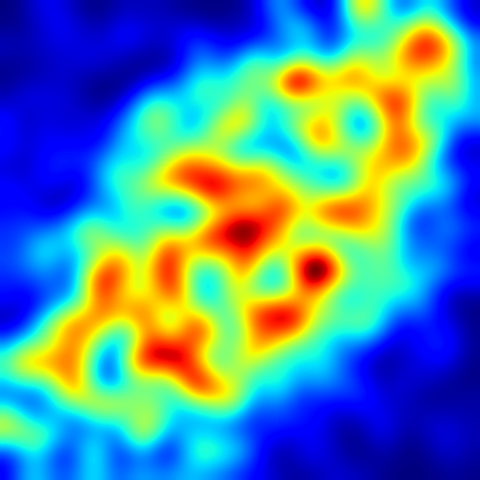}}
\fbox{\includegraphics[width=\heatmapwidth]{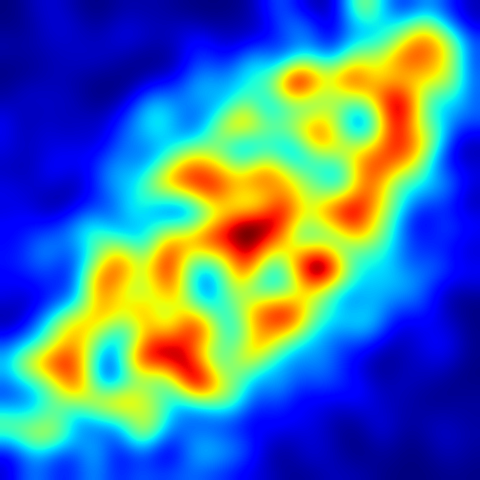}}
\caption{
Snapshots of the sample heatmap for \uniformguild (top) and IS (bottom) on the \forestenv environment.
Regions that have been sampled more densely are more red.
\uniformguild densifies around the optimal path more quickly than IS.
}
\label{fig:filmstrip}
\setlength{\fboxrule}{0.2pt}
\setlength{\fboxsep}{3pt}
\vspace{-1\baselineskip}
\end{figure}

\begin{table*}[t]
\centering
\renewcommand{\arraystretch}{1.1}
\begin{tabular}{@{}lcccc@{}}\toprule
& \texttt{IS (Baseline)} & \uniformguild & \greedyguild & \banditguild \\ \midrule
\forestenv        & $(2414, 2814)$ & $(1714, 1914)$ & ${\color{red} (1414, 1614)}$ & ${\color{red} (1410, 1690)}$ \\
\twowallforestenv & $(2614, 3414)$ & ${\color{red} (1814, 2414)}$ & ${\color{red} (1614, 2294)}$ & ${\color{red} (1830, 2310)}$ \\
\lforestenv       & $(3930, 6110)$ & ${\color{red} (1930, 2590)}$ & $(3410, 3990)$ & ${\color{red} (2210, 2410)}$ \\
\carenv           & $(2025, 2525)$ & ${\color{red} (1525, 1825)}$ & ${\color{red} (1525, 1725)}$ & ${\color{red} (1625, 1825)}$ \\
\herbenv          & $(4525, 5605)$ & ${\color{red} (1325, 1725)}$ & ${\color{red} (1225, 1605)}$ & ${\color{red} (1225, 1805)}$ \\
\bottomrule
\end{tabular}
\caption{
Median Sample Efficiency for converging to the optimal path cost,
with nonparametric 95\% confidence interval.
The best performing \BeaconSelector{} on each planning problem is highlighted.
}
\label{tab:sample-efficiency}
\end{table*}

\begin{figure*}[t]
\centering
\begin{subfigure}{0.205\linewidth}
  \includegraphics[width=\linewidth]{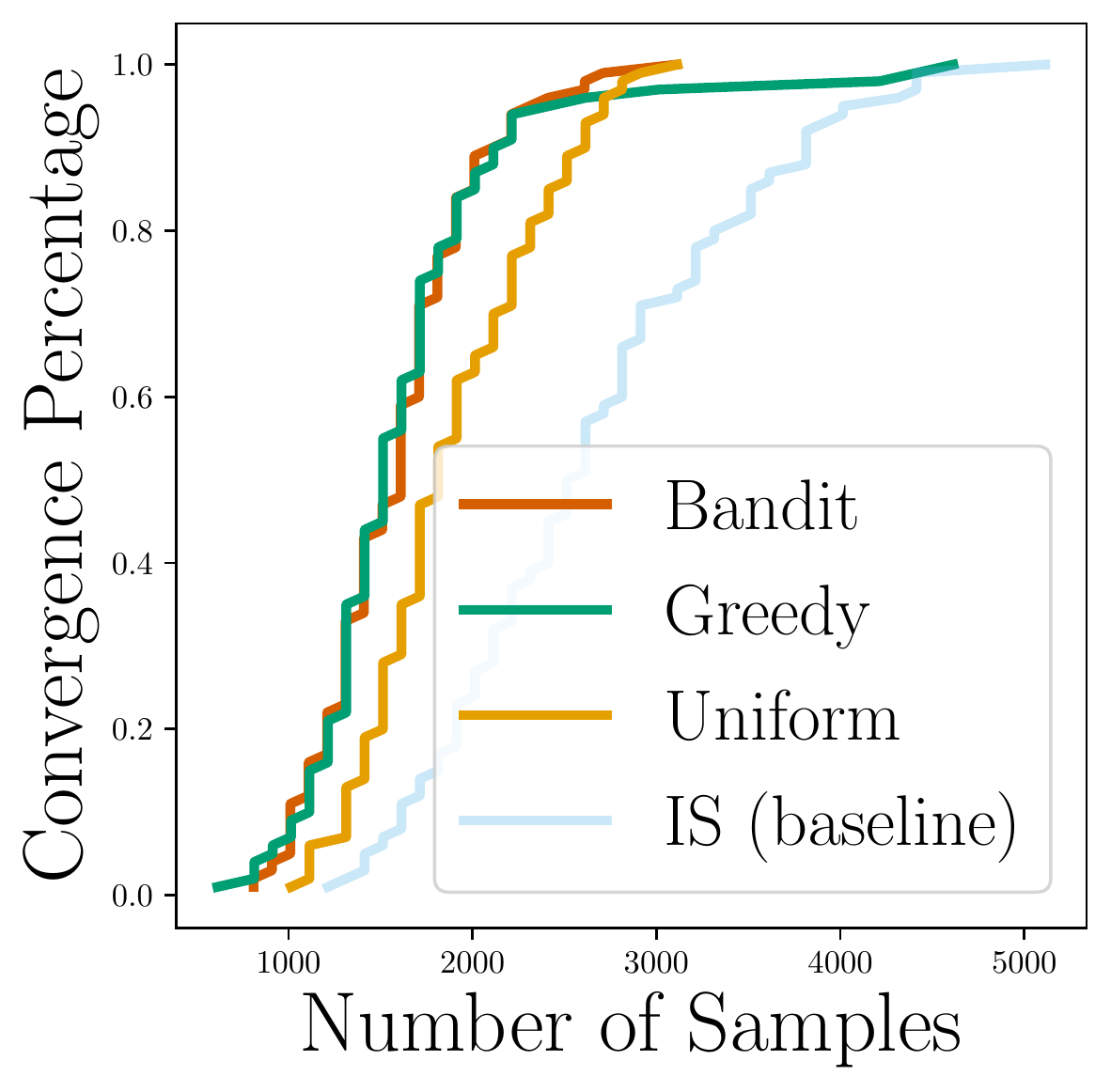}
  \caption{\forestenv}
  \label{fig:success_forest_hard}
\end{subfigure}
\begin{subfigure}{0.19\linewidth}
\includegraphics[width=\linewidth]{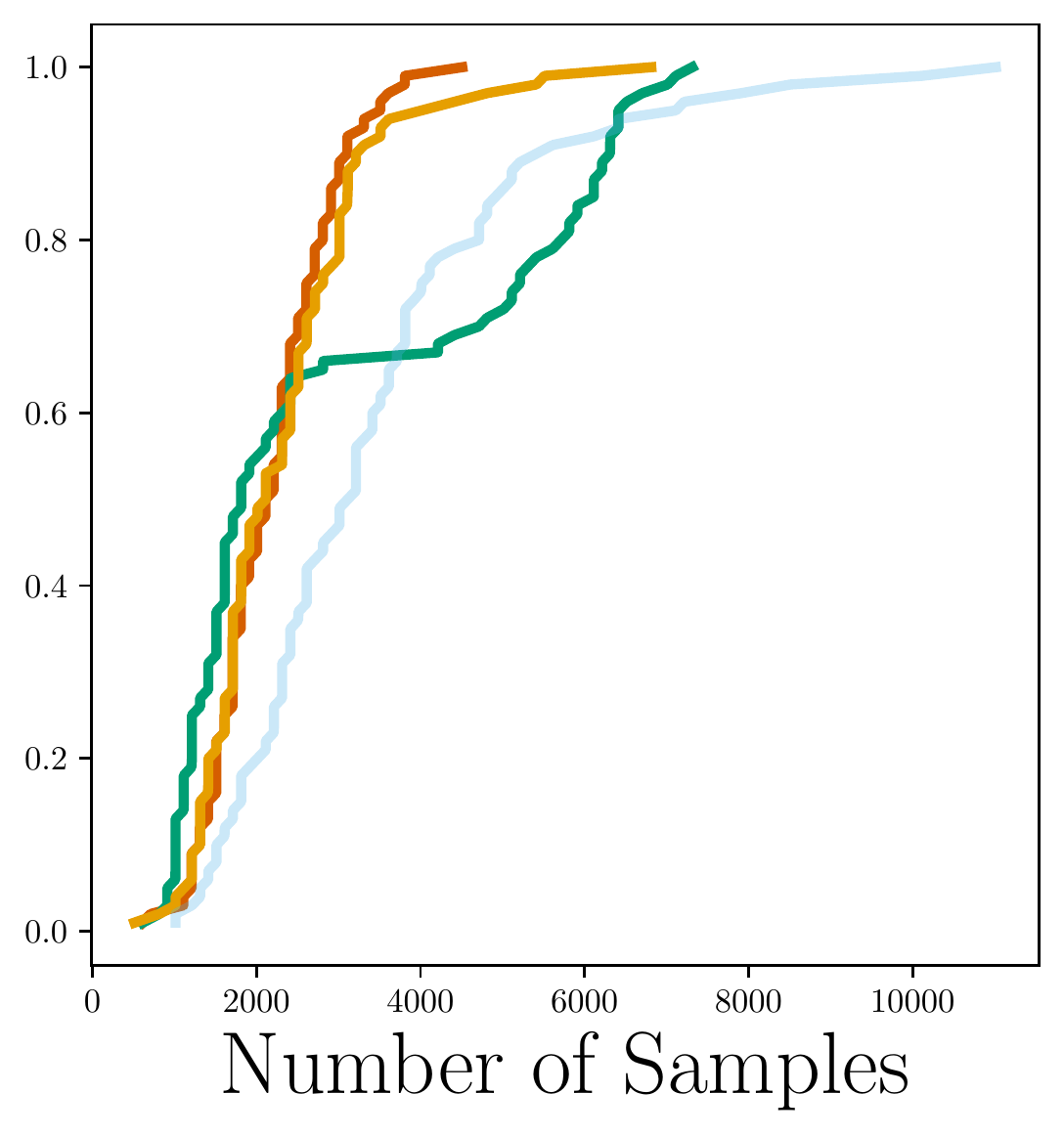}
  \caption{\twowallforestenv}
  \label{fig:success_two_wall_forest}
\end{subfigure}
\begin{subfigure}{0.2\linewidth}
\includegraphics[width=\linewidth]{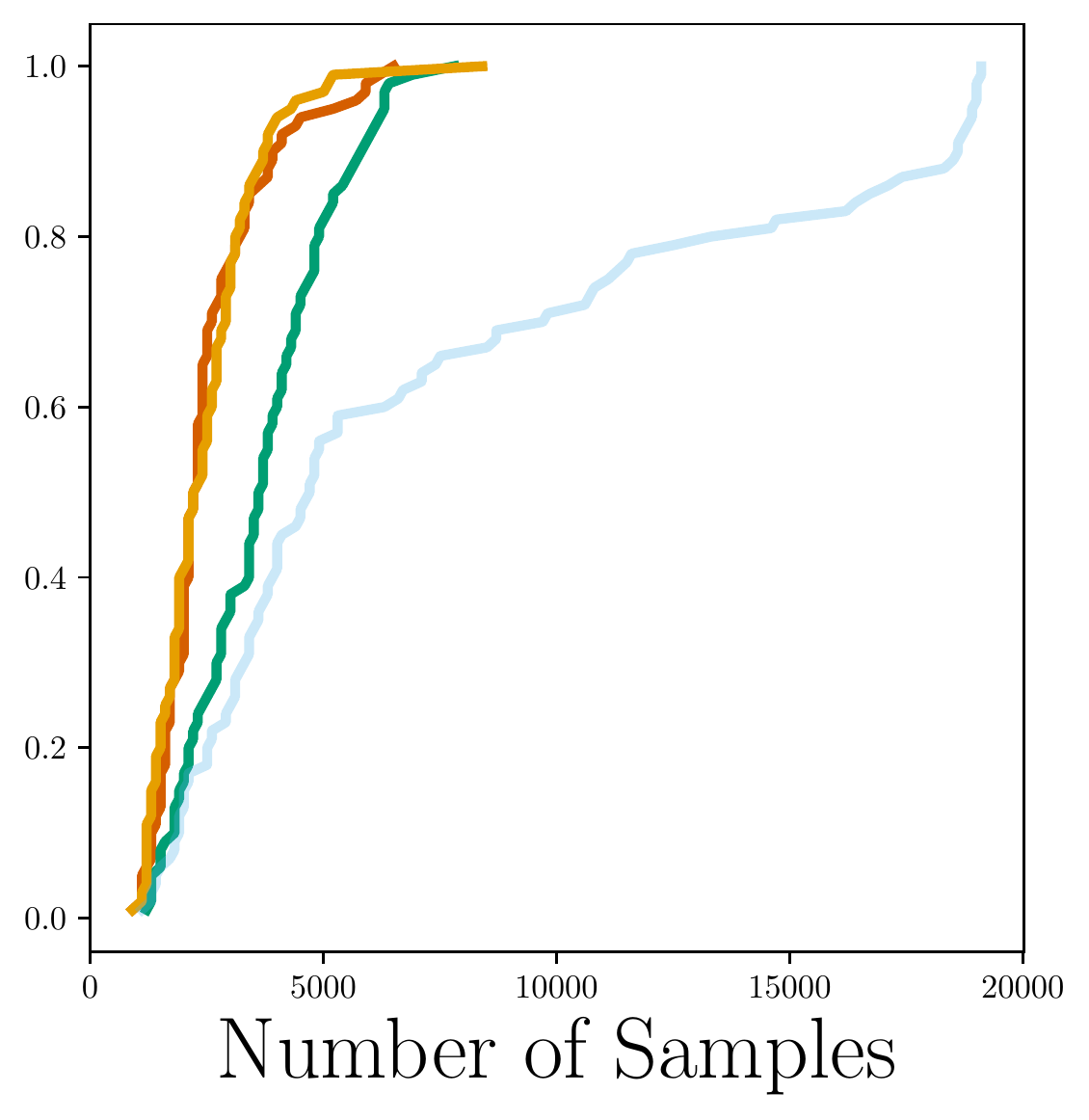}
  \caption{\lforestenv}
  \label{fig:success_l_forest}
\end{subfigure}
\begin{subfigure}{0.19\linewidth}
\includegraphics[width=\linewidth]{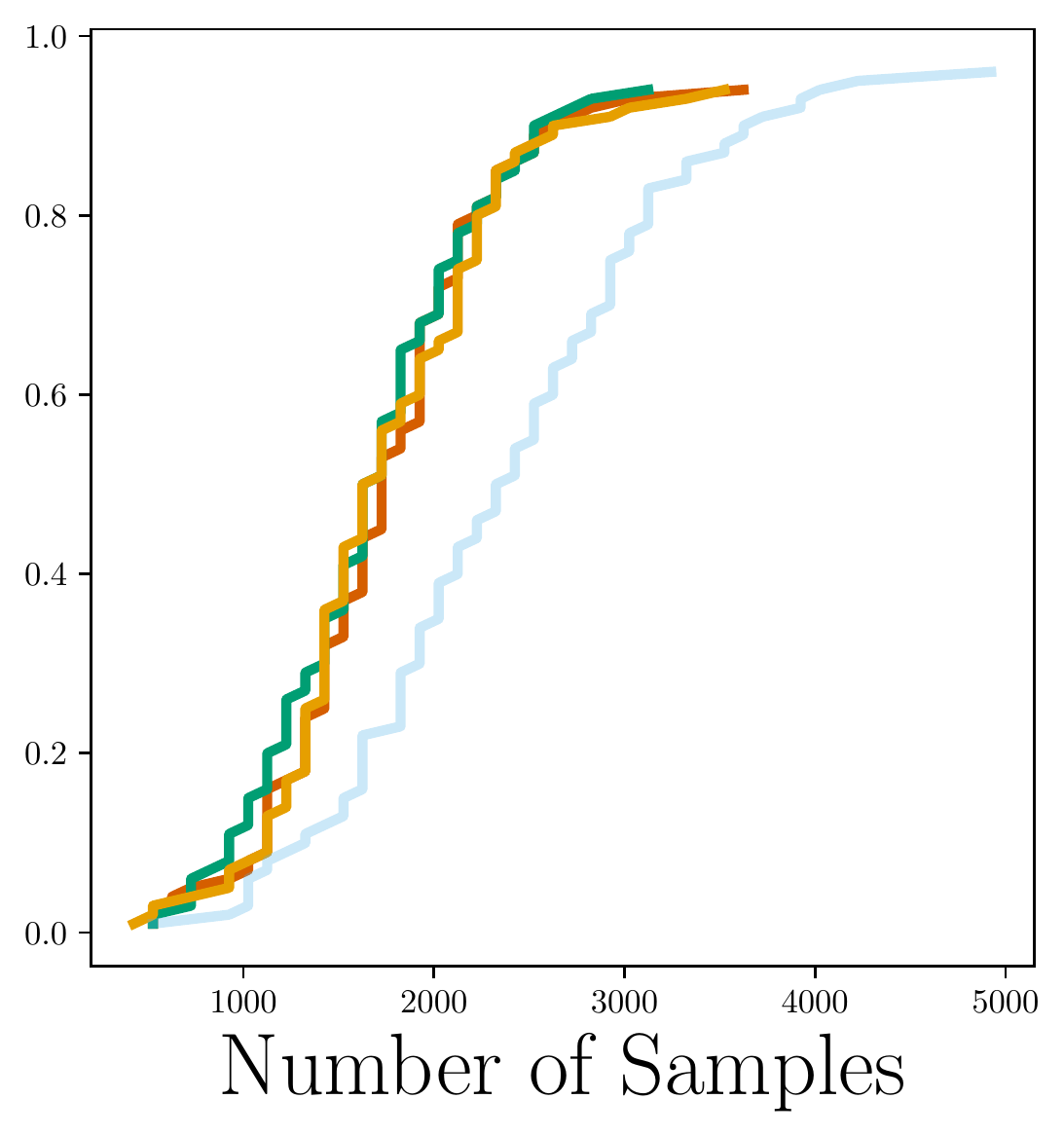}
  \caption{\carenv}
  \label{fig:success_ompl}
\end{subfigure}
\begin{subfigure}{0.19\linewidth}
\includegraphics[width=\linewidth]{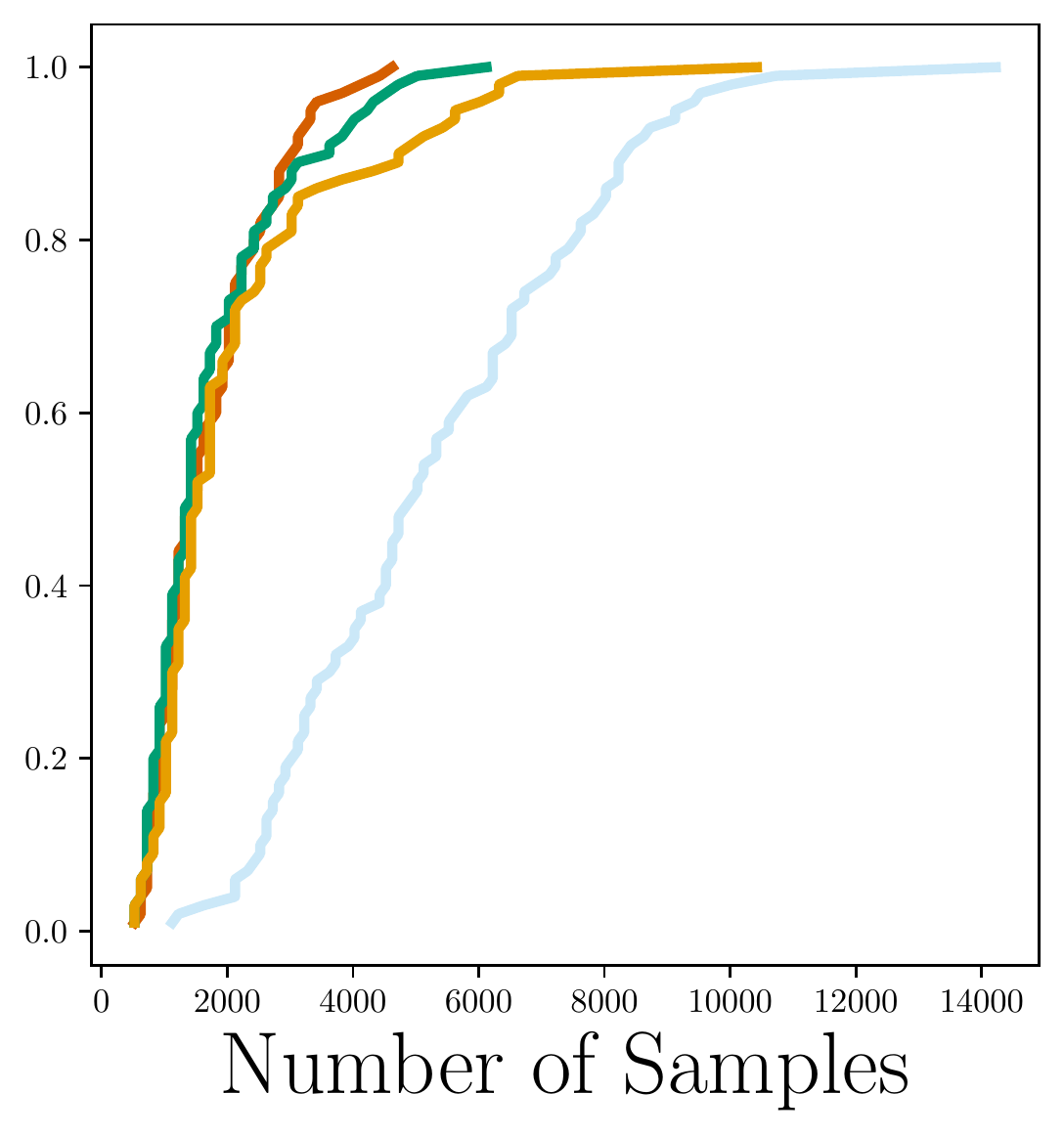}
  \caption{\herbenv}
  \label{fig:success_herb_shelf}
\end{subfigure}
\caption{
Convergence Percentage across 100 trials.
For most environments, sampling from the IS results in a worse Convergence Percentage than \guild for a given sample budget.
}
\label{fig:success_all}
\end{figure*}

\begin{figure*}[t]
\centering
\begin{subfigure}{0.205\linewidth}
  \includegraphics[width=\linewidth]{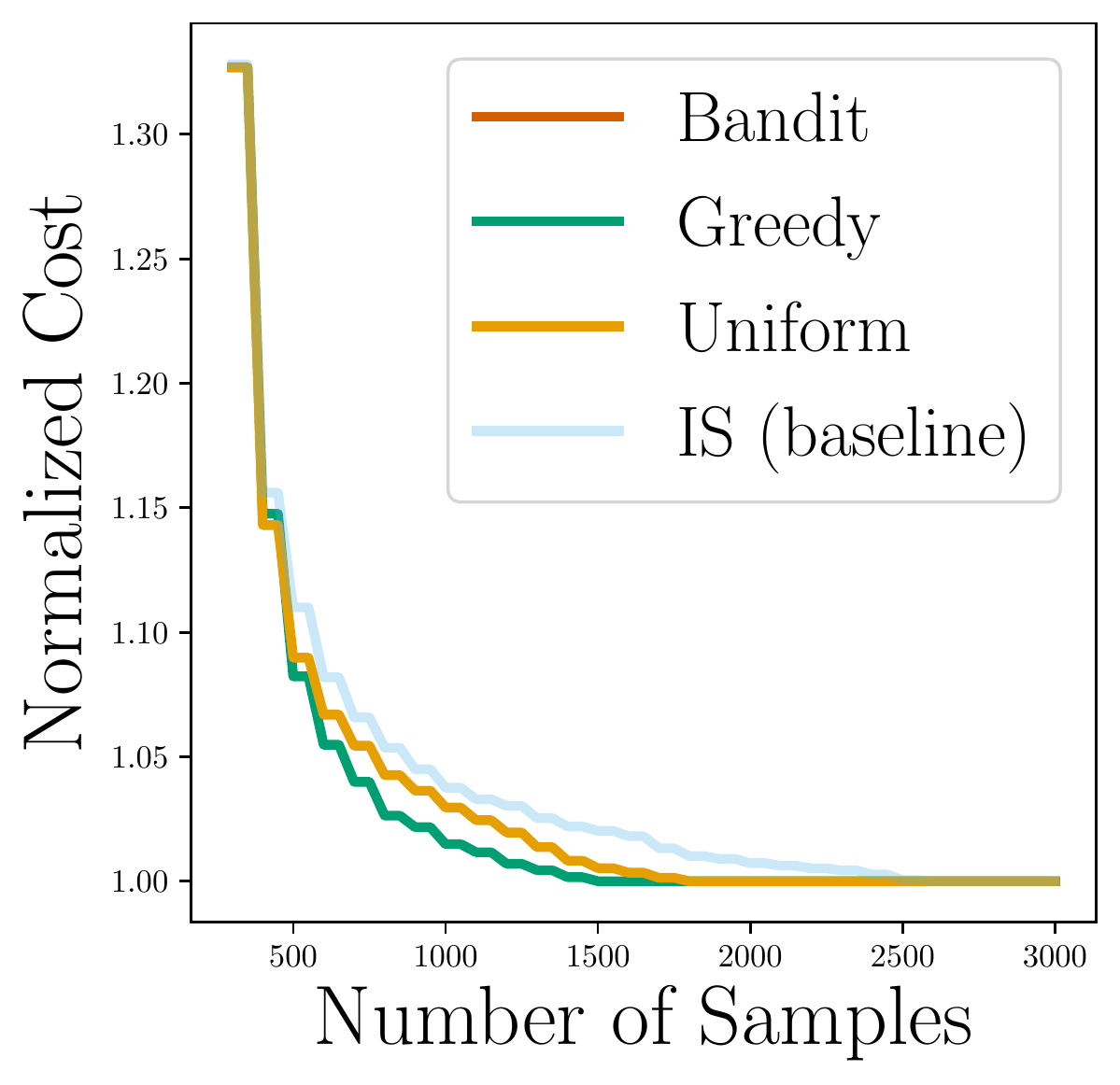}
  \caption{\forestenv}
  \label{fig:cost_forest_hard}
\end{subfigure}
\begin{subfigure}{0.19\linewidth}
\includegraphics[width=\linewidth]{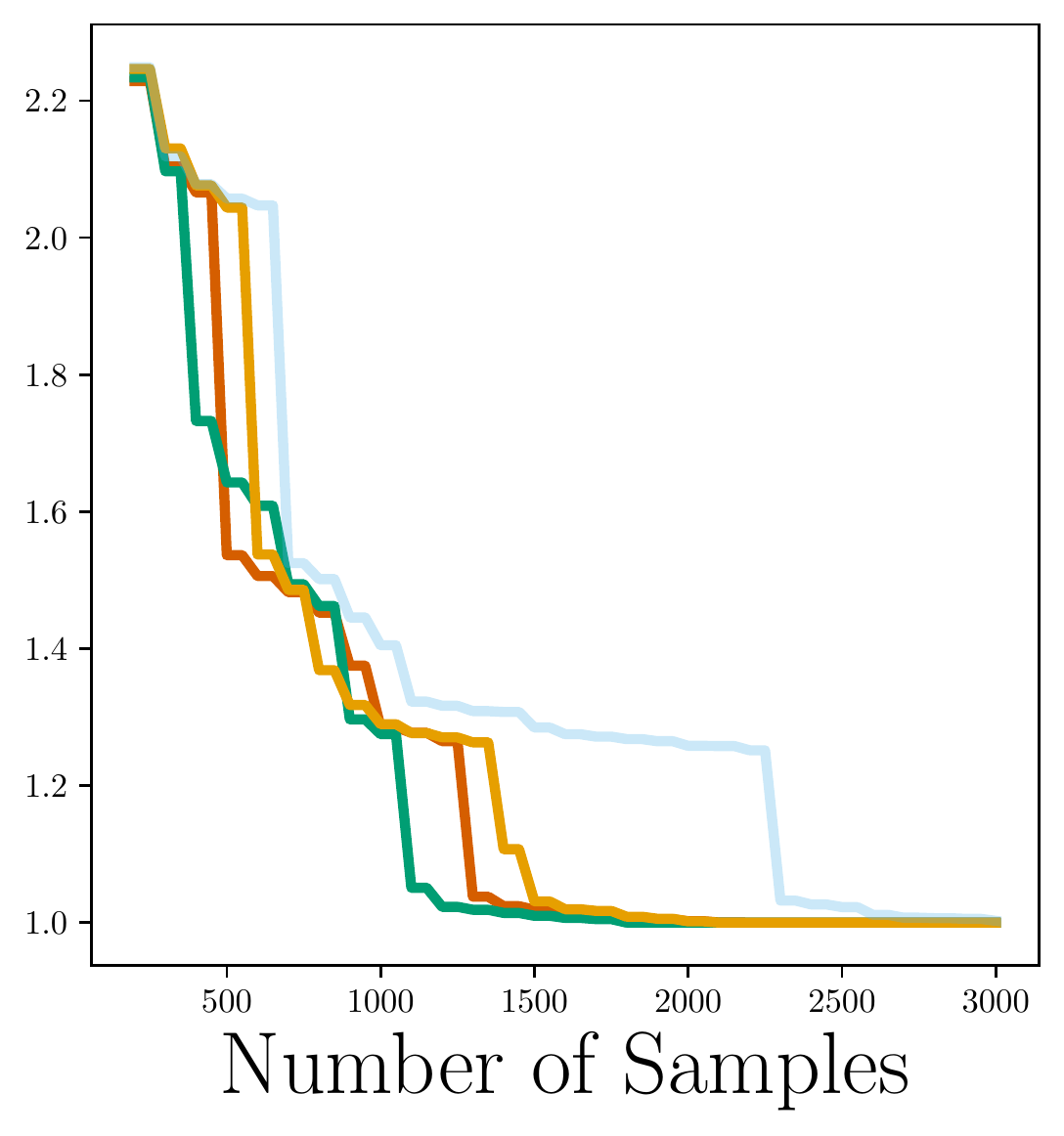}
  \caption{\twowallforestenv}
  \label{fig:cost_two_wall_forest}
\end{subfigure}
\begin{subfigure}{0.19\linewidth}
\includegraphics[width=\linewidth]{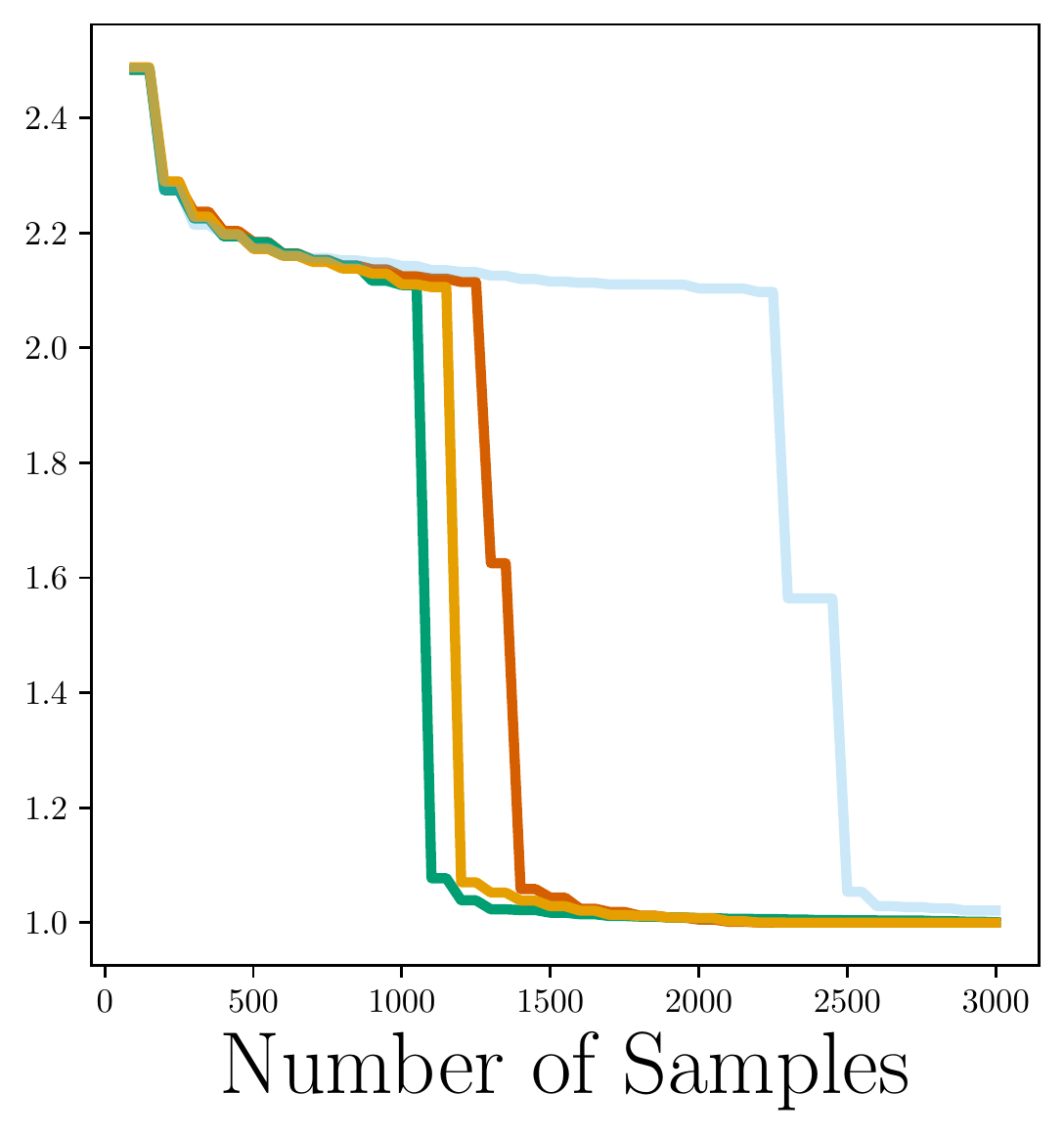}
  \caption{\lforestenv}
  \label{fig:cost_L_forest_hard}
\end{subfigure}
\begin{subfigure}{0.19\linewidth}
\includegraphics[width=\linewidth]{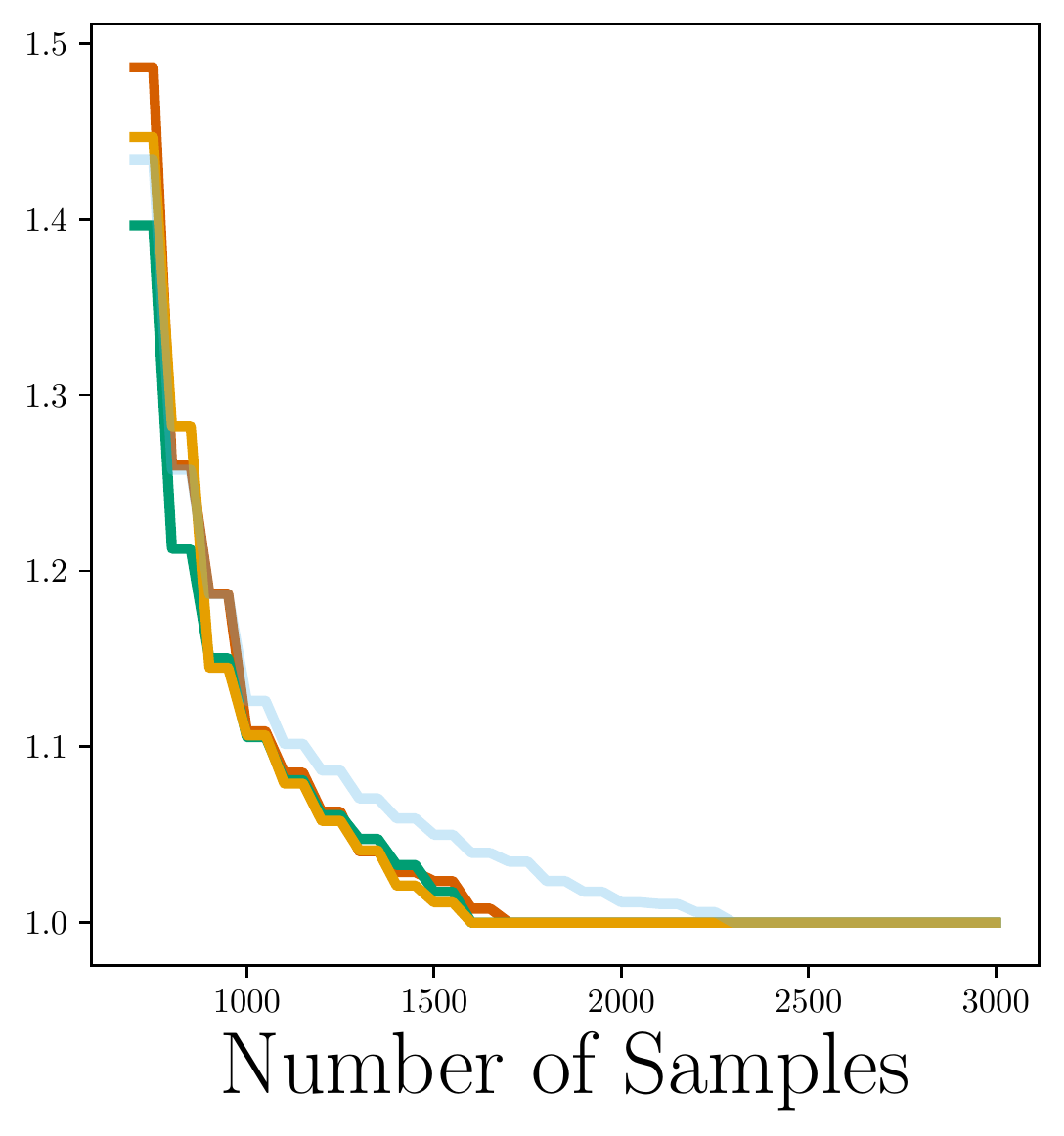}
  \caption{\carenv}
  \label{fig:cost_ompl}
\end{subfigure}
\begin{subfigure}{0.19\linewidth}
\includegraphics[width=\linewidth]{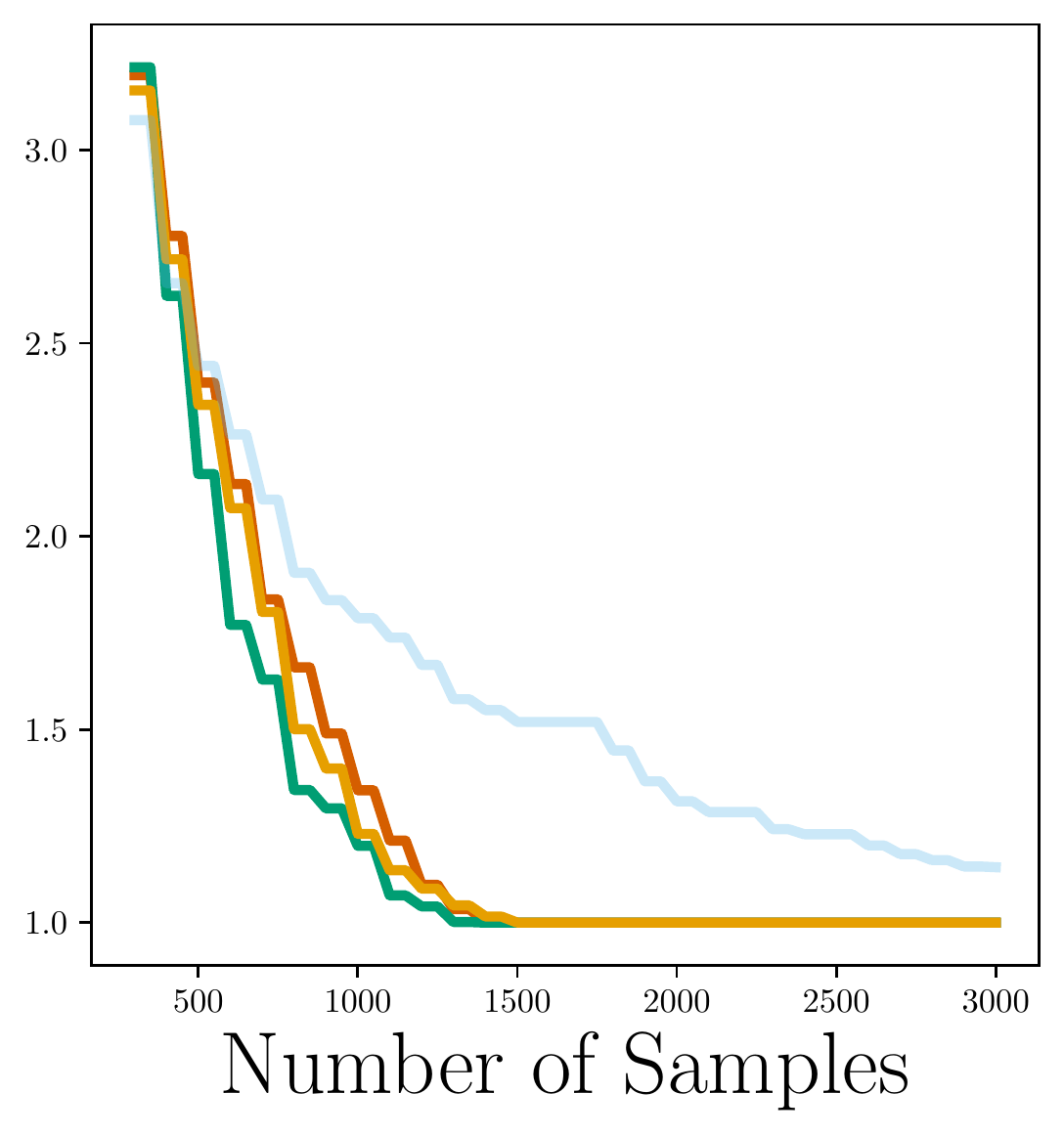}
  \caption{\herbenv}
  \label{fig:cost_herb_shelf}
\end{subfigure}
\caption{
Normalized Path Cost, median across 100 trials.
Sampling with \guild reaches a near-optimal path length more quickly than sampling from the IS,
with significant improvement in all but the easiest planning domain.
}
\label{fig:cost_all}
\end{figure*}

\subsection{Results}
\label{sec:results}

To test these hypotheses,
we implement the \BeaconSelector{}s described in \Cref{sec:guild-tree}.
We construct the set of beacons $\beaconSet{}$ by sampling states from a low-discrepancy sequence \cite{Halton64}.
We run 100 random trials for each pair of algorithm and environment.

First, we render a heatmap to visualize snapshots of the sampling distribution over time on the simple \forestenv environment (\Cref{fig:filmstrip}).
Qualitatively, \uniformguild focuses sampling around the eventual optimal path much more quickly than IS,
which we expect to yield improved Sample Efficiency and faster convergence.

\Cref{tab:sample-efficiency} reports a nonparametric 95\% confidence interval on the median Sample Efficiency for each instantiation of \guild.
We find that regardless of the beacon selector, \guild focuses sampling more efficiently than sampling from the IS, supporting \ref{hyp:sample-efficiency}.
In general, all three \guild selectors achieve comparable sample efficiency,
suggesting that the key to their success is sampling from Local Subsets.
However, the \banditguild selector most consistently ranks among the best performing \BeaconSelector{}s,
while \uniformguild and \greedyguild were each marginally less efficient on one environment.

To understand convergence across the random trials,
we plot the Convergence Percentage as a function of the number of samples (\Cref{fig:success_all}).
To support \ref{hyp:success-ratio},
we would expect the IS baseline curve to remain below the \guild curves.
We find this to be the case:
on most environments, sampling from the IS results in a lower Convergence Percentage for a fixed sample budget.
However, on a third of the \twowallforestenv trials,
the \greedyguild heuristic causes over-sampling in regions that ultimately do not yield the optimal path.

While \ref{hyp:sample-efficiency} shows that \guild ultimately converges faster to the optimal cost,
we would also like to characterize the rate of convergence.
For each planning problem,
we plot the median Normalized Path Cost across the trials to understand
how path length is reduced over time
(\Cref{fig:cost_all}).
Sharp drops in path length (e.g., \Cref{fig:cost_L_forest_hard})
highlight the discrepancy between
high-cost homotopies that are easy to sample (navigating around large obstacles)
and near-optimal homotopies that are difficult to sample (crossing narrow passages through obstacles).
Steady decreases in path length (e.g., \Cref{fig:cost_forest_hard}) show planning problems where
sampling is not stuck in high-cost homotopies and can discover low-cost homotopies more easily.

\ref{hyp:path-cost} is supported across all environments:
sampling with \guild consistently yields a lower cost than sampling with IS.
In planning problems with low-cost homotopies,
\guild automatically focuses sampling to deliver paths through these narrow passages.
With the \lforestenv environment,
\guild instances identify the narrow passage around 1100-1200 samples
and quickly optimize within homotopies crossing the passage.
By contrast, the IS baseline requires nearly double the samples to find the passage.
The \twowallforestenv environment shows a similar trend,
although with additional intermediate-cost homotopies that only traverse one of the two narrow passages
(\Cref{fig:cost_two_wall_forest}).

The \carenv and \herbenv demonstrate the anytime performance of \guild on
environments with less distinct homotopy costs.
As a result, all sampling schemes steadily improve path cost,
similar to \forestenv.
However, in \herbenv, \guild instances have a much steeper improvement,
suggesting that \guild may yield more significant sample efficiency improvements in higher dimensions.

\section{Conclusion}
\label{sec:Conclusion}

\guild is a new framework for incremental densification with a simple insight:
even when the planner fails to discover a shorter path,
the search tree still contains valuable information that can immediately improve the densification strategy.
\guild uses the search tree to select a \emph{beacon},
a vertex in the search tree that decomposes the original sampling/planning problem into two smaller subproblems.
Improving the cost-to-come for any beacon allows \guild to shrink and expand the Local Subsets that it samples from.
Similar to the Informed Set that \guild builds upon,
Local Subsets can be easily (and efficiently) incorporated into any sampling-based optimal motion planning algorithm.

We demonstrate that even simple beacon selectors,
such as \uniformguild,
can dramatically accelerate the convergence rate relative to the Informed Set densification baseline.
We also propose a \banditguild selector using EXP3,
an adversarial bandit algorithm that consistently ranks among the best beacon selectors across all the planning problems we considered.
In particular, \guild excels in domains with difficult-to-sample homotopy classes and high-dimensional planning problems.

In this work,
we have primarily considered either heuristic beacon selectors (\greedyguild) or beacon selectors that learn from experience online (\banditguild).
Better heuristics for selecting beacons may exist,
including ones that build on prior work in identifying and sampling bottleneck points.
We believe that experience in the form of large planning problem datasets may also provide valuable information that will guide beacon selection to sample even more efficiently.

\bibliographystyle{unsrt}
\bibliography{references}

\begin{thebibliography}{10}

\bibitem{KF11}
Sertac Karaman and Emilio Frazzoli.
\newblock Sampling-based algorithms for optimal motion planning.
\newblock {\em I. J. Robotics Res.}, 30(7):846--894, 2011.

\bibitem{karaman10Incremental}
Sertac Karaman and Emilio Frazzoli.
\newblock Incremental sampling-based algorithms for optimal motion planning.
\newblock In {\em RSS}, 2010.

\bibitem{ArslanT13}
Oktay Arslan and Panagiotis Tsiotras.
\newblock Use of relaxation methods in sampling-based algorithms for optimal
  motion planning.
\newblock In {\em ICRA}, pages 2421--2428, 2013.

\bibitem{GSB15}
Jonathan~D. Gammell, Siddhartha~S. Srinivasa, and Timothy~D. Barfoot.
\newblock {Batch Informed Trees ({BIT}*): Sampling-based optimal planning via
  the heuristically guided search of implicit random geometric graphs}.
\newblock In {\em ICRA}, pages 3067--3074, 2015.

\bibitem{StrubG20ABIT}
Marlin~P. Strub and Jonathan~D. Gammell.
\newblock {Advanced BIT* (ABIT*): Sampling-Based Planning with Advanced
  Graph-Search Techniques}.
\newblock In {\em ICRA}, pages 130--136. {IEEE}, 2020.

\bibitem{StrubG20AIT}
Marlin~P. Strub and Jonathan~D. Gammell.
\newblock {Adaptively Informed Trees (AIT*): Fast Asymptotically Optimal Path
  Planning through Adaptive Heuristics}.
\newblock In {\em ICRA}, pages 3191--3198. {IEEE}, 2020.

\bibitem{gammell2014informed}
Jonathan~D. Gammell, Siddhartha~S. Srinivasa, and Timothy~D. Barfoot.
\newblock {Informed RRT*: Optimal sampling-based path planning focused via
  direct sampling of an admissible ellipsoidal heuristic}.
\newblock In {\em 2014 IEEE/RSJ International Conference on Intelligent Robots
  and Systems}, pages 2997--3004. IEEE, 2014.

\bibitem{kavraki96prm}
Lydia~E. Kavraki, Petr Svestka, Jean-Claude Latombe, and Mark~H. Overmars.
\newblock Probabilistic roadmaps for path planning in high-dimensional
  configuration spaces.
\newblock {\em {IEEE} Trans. Robotics and Automation}, 12(4):566--580, 1996.

\bibitem{lavallek01}
Steven~M. LaValle and James J.~Kuffner Jr.
\newblock Randomized kinodynamic planning.
\newblock {\em Int. J. Robotics Res.}, 20(5):378--400, 2001.

\bibitem{kuffner2000rrt}
James J.~Kuffner Jr. and Steven~M. LaValle.
\newblock {RRT-Connect: An Efficient Approach to Single-Query Path Planning}.
\newblock In {\em ICRA}, volume~2, pages 995--1001. IEEE, 2000.

\bibitem{hsuEST}
David Hsu, Jean{-}Claude Latombe, and Rajeev Motwani.
\newblock Path planning in expansive configuration spaces.
\newblock In {\em ICRA}, pages 2719--2726. {IEEE}, 1997.

\bibitem{JSCP15}
Lucas Janson, Edward Schmerling, Ashley~A. Clark, and Marco Pavone.
\newblock Fast marching tree: {A} fast marching sampling-based method for
  optimal motion planning in many dimensions.
\newblock {\em I. J. Robotics Res.}, 34(7):883--921, 2015.

\bibitem{wilmarth1999maprm}
Steven~A. Wilmarth, Nancy~M. Amato, and Peter~F. Stiller.
\newblock {MAPRM: A probabilistic roadmap planner with sampling on the medial
  axis of the free space}.
\newblock In {\em ICRA}, 1999.

\bibitem{holleman2000framework}
Christopher Holleman and Lydia~E. Kavraki.
\newblock {A framework for using the workspace medial axis in PRM planners}.
\newblock In {\em ICRA}, 2000.

\bibitem{hsu2003bridge}
David Hsu, Tingting Jiang, John~H. Reif, and Zheng Sun.
\newblock The bridge test for sampling narrow passages with probabilistic
  roadmap planners.
\newblock In {\em ICRA}, 2003.

\bibitem{kurniawati2004workspace}
Hanna Kurniawati and David Hsu.
\newblock Workspace importance sampling for probabilistic roadmap planning.
\newblock In {\em IROS}, volume~2, pages 1618--1623. IEEE, 2004.

\bibitem{yang2004adapting}
Y.~Yang and O.~Brock.
\newblock Adapting the sampling distribution in prm planners based on an
  approximated medial axis.
\newblock In {\em ICRA}, 2004.

\bibitem{van2005using}
J.~P. Van~den Berg and M.~H. Overmars.
\newblock Using workspace information as a guide to non-uniform sampling in
  probabilistic roadmap planners.
\newblock {\em IJRR}, 2005.

\bibitem{DennyGTA14}
Jory Denny, Evan Greco, Shawna~L. Thomas, and Nancy~M. Amato.
\newblock {MARRT: Medial Axis biased rapidly-exploring random trees}.
\newblock In {\em ICRA}, pages 90--97. {IEEE}, 2014.

\bibitem{amato1998obprm}
Nancy~M. Amato, Bayazit, O.~Burchan Bayazit, Lucia~K. Dale, Christopher Jones,
  and Daniel Vallejo.
\newblock {OBPRM: An obstacle-based PRM for 3D workspaces}.
\newblock 1998.

\bibitem{boor1999gaussian}
Val{\'{e}}rie Boor, Mark~H. Overmars, and A.~Frank van~der Stappen.
\newblock {The Gaussian Sampling Strategy for Probabilistic Roadmap Planners}.
\newblock In {\em ICRA}, pages 1018--1023, 1999.

\bibitem{amatoOBRRT}
Samuel Rodr{\'{\i}}guez, Xinyu Tang, Jyh{-}Ming Lien, and Nancy~M. Amato.
\newblock An obstacle-based rapidly-exploring random tree.
\newblock In {\em ICRA}, pages 895--900, 2006.

\bibitem{doersch2016tutorial}
Carl Doersch.
\newblock Tutorial on variational autoencoders.
\newblock {\em arXiv preprint arXiv:1606.05908}, 2016.

\bibitem{ichter2017learning}
Brian Ichter, James Harrison, and Marco Pavone.
\newblock Learning sampling distributions for robot motion planning.
\newblock In {\em ICRA}, 2018.

\bibitem{kumar2019lego}
Rahul Kumar, Aditya Mandalika, Sanjiban Choudhury, and Siddhartha~S. Srinivasa.
\newblock {LEGO: Leveraging Experience in Roadmap Generation for Sampling-Based
  Planning}.
\newblock In {\em IROS}, pages 1488--1495. {IEEE}, 2019.

\bibitem{IchterSLF20}
Brian Ichter, Edward Schmerling, Tsang{-}Wei~Edward Lee, and Aleksandra Faust.
\newblock Learned critical probabilistic roadmaps for robotic motion planning.
\newblock In {\em ICRA}, pages 9535--9541. {IEEE}, 2020.

\bibitem{chamzas2019primitives}
Constantinos Chamzas, Anshumali Shrivastava, and Lydia~E. Kavraki.
\newblock {Using Local Experiences for Global Motion Planning}.
\newblock In {\em ICRA}, pages 8606--8612, 2019.

\bibitem{chamzas3D}
Constantinos Chamzas, Zachary~K. Kingston, Carlos Quintero{-}Pe{\~{n}}a,
  Anshumali Shrivastava, and Lydia~E. Kavraki.
\newblock {Learning Sampling Distributions Using Local 3D Workspace
  Decompositions for Motion Planning in High Dimensions}.
\newblock {\em CoRR}, abs/2010.15335, 2020.

\bibitem{arbaazGNN}
Arbaaz Khan, Alejandro Ribeiro, Vijay Kumar, and Anthony~G. Francis.
\newblock Graph neural networks for motion planning.
\newblock {\em CoRR}, abs/2006.06248, 2020.

\bibitem{DennyA12}
Jory Denny and Nancy~M. Amato.
\newblock Toggle {PRM:} {A} coordinated mapping of c-free and c-obstacle in
  arbitrary dimension.
\newblock In {\em WAFR}, volume~86, pages 297--312, 2012.

\bibitem{diankov2007randomized}
Rosen Diankov and James Kuffner.
\newblock Randomized statistical path planning.
\newblock In {\em IROS}, 2007.

\bibitem{JailletCS10}
Leonard Jaillet, Juan Cort{\'{e}}s, and Thierry Sim{\'{e}}on.
\newblock {Sampling-Based Path Planning on Configuration-Space Costmaps}.
\newblock {\em {IEEE} Trans. Robotics}, 26(4):635--646, 2010.

\bibitem{HauerT17}
Florian Hauer and Panagiotis Tsiotras.
\newblock Deformable rapidly-exploring random trees.
\newblock In {\em RSS}, 2017.

\bibitem{PhillipsBK04}
Jeff~M. Phillips, Nazareth Bedrossian, and Lydia~E. Kavraki.
\newblock {Guided Expansive Spaces Trees: a Search Strategy for Motion- and
  Cost-constrained State Spaces}.
\newblock In {\em ICRA}, pages 3968--3973. {IEEE}, 2004.

\bibitem{ArslanT15RR}
Oktay Arslan and Panagiotis Tsiotras.
\newblock Dynamic programming guided exploration for sampling-based motion
  planning algorithms.
\newblock In {\em ICRA}, pages 4819--4826. {IEEE}, 2015.

\bibitem{ArslanT15ML}
Oktay Arslan and Panagiotis Tsiotras.
\newblock Machine learning guided exploration for sampling-based motion
  planning algorithms.
\newblock In {\em IROS}, pages 2646--2652. {IEEE}, 2015.

\bibitem{JoshiT20}
Sagar~Suhas Joshi and Panagiotis Tsiotras.
\newblock {Relevant Region Exploration On General Cost-maps For Sampling-Based
  Motion Planning}.
\newblock In {\em IROS}, pages 6689--6695. {IEEE}, 2020.

\bibitem{GSB20}
Jonathan~D. Gammell, Timothy~D. Barfoot, and Siddhartha~S. Srinivasa.
\newblock Batch informed trees (bit*): Informed asymptotically optimal anytime
  search.
\newblock {\em IJRR}, 39(5), 2020.

\bibitem{auer2002}
Peter Auer, Nicolo Cesa-Bianchi, Yoav Freund, and Robert~E Schapire.
\newblock The nonstochastic multiarmed bandit problem.
\newblock {\em SIAM journal on computing}, 32(1):48--77, 2002.

\bibitem{ompl}
Ioan~Alexandru Sucan, Mark Moll, and Lydia~E. Kavraki.
\newblock The open motion planning library.
\newblock {\em IEEE Robotics Automation Magazine}, 19(4):72--82, Dec 2012.

\bibitem{srinivasa2010herb}
Siddhartha~S Srinivasa, Dave Ferguson, Casey~J Helfrich, Dmitry Berenson,
  Alvaro Collet, Rosen Diankov, Garratt Gallagher, Geoffrey Hollinger, James
  Kuffner, and Michael~Vande Weghe.
\newblock Herb: a home exploring robotic butler.
\newblock {\em Autonomous Robots}, 28(1):5--20, 2010.

\bibitem{Halton64}
J.~H. Halton and G.~B. Smith.
\newblock Algorithm 247: Radical-inverse quasi-random point sequence.
\newblock {\em Commun. ACM}, 7(12):701--702, December 1964.

\end{thebibliography}

\end{document}